\documentclass[conference]{IEEEtran}
\IEEEoverridecommandlockouts

\usepackage{cite}
\usepackage{amsmath,amssymb,amsfonts}
\usepackage{algorithmic}
\usepackage{graphicx}
\usepackage{textcomp}
\usepackage{xcolor}
\usepackage{booktabs}
\usepackage{array}
\usepackage{tabularx}
\usepackage{enumitem}
\usepackage{tikz}
\usetikzlibrary{arrows.meta,positioning,fit,backgrounds,calc}
\usepackage[hyphens]{url}
\usepackage[hidelinks]{hyperref}
\def\BibTeX{{\rm B\kern-.05em{\sc i\kern-.025em b}\kern-.08em
    T\kern-.1667em\lower.7ex\hbox{E}\kern-.125emX}}

\definecolor{slateborder}{RGB}{71,85,105}
\definecolor{slatefill}{RGB}{248,250,252}
\definecolor{blueborder}{RGB}{37,99,235}
\definecolor{bluefill}{RGB}{239,246,255}
\definecolor{tealborder}{RGB}{13,148,136}
\definecolor{tealfill}{RGB}{240,253,250}
\definecolor{amberborder}{RGB}{217,119,6}
\definecolor{amberfill}{RGB}{255,251,235}
\definecolor{fmbadge}{RGB}{220,38,38}

\begin{document}
\bstctlcite{IEEEexample:BSTcontrol}

\title{Harnesses That Improve Themselves: A Unifying Framework for Self-Improving Agent Harnesses}

\author{\IEEEauthorblockN{Deep Shah}
\IEEEauthorblockA{\textit{Google}\\
shahdeep@google.com}
\and
\IEEEauthorblockN{Nehal Kathrotia}
\IEEEauthorblockA{\textit{Google}\\
nehalk@google.com}}

\maketitle

\begin{abstract}
Whether a foundation model succeeds as an autonomous agent depends as much on its runtime software as on its weights. This runtime, the \emph{agent harness}, manages context and memory, steers multi-turn control, executes tool actions in sandboxes, and verifies work before stopping. Because models differ in how they use tools and how they fail, a static harness tuned for one model or task often stalls, wastes tokens, or regresses on another. This bottleneck has driven a shift from prompt tuning to \emph{self-improving agent harnesses}: systems that inspect their own execution traces, separate runtime bugs from model limits, and rewrite their own code. We present a unifying architectural and design framework for self-improving agent harnesses. The framework pairs four editable runtime components (context and state, agent control, action interfaces, and verification and recovery) with a four-stage loop that diagnoses failures, proposes code patches, validates candidates, and deploys updates. Using this framework, we show how optimizers trace errors across long trajectories and discover reusable runtime mechanisms such as pre-stop verifiers, loop breakers, and persistent memory. We further show why full-harness rewrites overfit or break across modules, and how modular updates with validation gates generalize across models while bounding token cost. To keep self-modification safe, we establish practical defenses against recurring failure modes including prompt bloat, benchmark gaming, and evaluator tampering. Finally, experiments on SWE-bench Lite validate this framework, showing that gated modular evolution outperforms prompt reflection and full-harness rewrites across models and repositories at a fraction of the token cost.
\end{abstract}

\begin{IEEEkeywords}
LLM agents, agent harness, harness engineering, automated harness optimization, recursive self-improvement, context engineering, agent evaluation
\end{IEEEkeywords}

\begin{figure*}[!t]
\centering
\resizebox{\textwidth}{!}{%
\begin{tikzpicture}[font=\sffamily\scriptsize, >=Stealth,
  panel/.style={draw=slateborder!45, fill=slatefill, rounded corners=5pt, line width=0.6pt},
  core/.style={draw=slateborder, fill=white, rounded corners=3pt, align=center, inner sep=3.0pt, line width=0.65pt, minimum height=0.76cm},
  comp/.style={draw=blueborder!75, fill=bluefill!70, rounded corners=3pt, align=center, inner sep=2.5pt, line width=0.6pt, text width=2.24cm, minimum height=0.92cm},
  stage/.style={draw=tealborder!80, fill=tealfill!75, rounded corners=3pt, align=center, inner sep=2.6pt, line width=0.6pt, text width=2.72cm, minimum height=0.92cm},
  flow/.style={->, line width=0.7pt, draw=slateborder!85},
  bridge/.style={->, line width=0.85pt}]

\draw[panel] (0,0) rectangle (8.90,4.55);
\node[anchor=north west, font=\sffamily\scriptsize\bfseries, text=slateborder] at (0.25,4.42) {Inner Loop: Interactive Agent Execution};

\node[core, text width=8.14cm] (model) at (4.45,3.56) {\textbf{Frozen Foundation Model}\\ \textcolor{slateborder}{reads curated context $\cdot$ emits reasoning, code \& tool calls}};

\node[comp] (c1) at (1.48,2.14) {\textbf{1. Context \& State}\\ memory, rules \&\\ compaction};
\node[comp, fill=white, draw=slateborder!75] (c2) at (4.45,2.14) {\textbf{2. Agent Control}\\ turn loop, retries\\ \& subagents};
\node[comp] (c3) at (7.42,2.14) {\textbf{3. Action Interface}\\ tools, shell ACI\\ \& sandbox};

\node[comp, draw=amberborder!85, fill=amberfill!85] (c4) at (1.48,0.64) {\textbf{4. Verification \&}\\ \textbf{Recovery}\\ \mbox{tests \& repair hooks}};
\node[core, text width=4.15cm, minimum height=0.92cm] (env) at (6.45,0.64) {\textbf{Live Environment}\\ \textcolor{slateborder}{\mbox{code repo\,$\cdot$\,terminal\,$\cdot$\,GUI\,$\cdot$\,web\,$\cdot$\,robot}}};

\draw[flow] (c1.north) -- (c1.north |- model.south) node[midway, right, font=\sffamily\tiny, text=slateborder] {context};
\draw[flow] (c2.north) -- (c2.north |- model.south) node[midway, right, font=\sffamily\tiny, text=slateborder] {steer};
\draw[flow] (c3.north |- model.south) -- (c3.north) node[midway, left, font=\sffamily\tiny, text=slateborder] {action};
\draw[<->, line width=0.7pt, draw=slateborder!85] (c3.south) -- (c3.south |- env.north) node[midway, left, font=\sffamily\tiny, text=slateborder] {execute / obs.};
\draw[flow, dashed] ([xshift=-0.50cm]c2.south) |- ([yshift=0.16cm]c4.east) node[pos=0.74, above, font=\sffamily\tiny, text=slateborder] {stop / check};
\draw[<->, line width=0.7pt, draw=slateborder!85] ([yshift=-0.16cm]c4.east) -- ([yshift=-0.16cm]env.west) node[midway, below, font=\sffamily\tiny, text=slateborder] {run tests};
\draw[flow] (c4.north) -- (c1.south) node[midway, left, font=\sffamily\tiny, text=slateborder] {repair / exit};
\draw[<->, line width=0.55pt, dashed, draw=slateborder!70] (c1.east) -- (c2.west);
\draw[<->, line width=0.55pt, dashed, draw=slateborder!70] (c2.east) -- (c3.west);

\draw[panel] (10.75,0) rectangle (18.15,4.55);
\node[anchor=north west, font=\sffamily\scriptsize\bfseries, text=tealborder!90!black] at (11.00,4.42) {Outer Loop: Closed-Loop Harness Self-Improvement};

\node[stage] (s1) at (12.48,3.18) {\textbf{1. Diagnose Failures}\\ cluster verifier errors\\ \& diff trajectories};
\node[stage] (s2) at (16.42,3.18) {\textbf{2. Propose Edits}\\ synthesize hook or\\ module code patch};
\node[stage] (s3) at (16.42,1.00) {\textbf{3. Validate \& Filter}\\ check runtime execution\\ \& held-out tasks};
\node[stage] (s4) at (12.48,1.00) {\textbf{4. Deploy or Route}\\ merge into default, save\\ to archive, or route};

\draw[flow] (s1.east) -- (s2.west) node[midway, above, font=\sffamily\tiny, text=slateborder] {root causes};
\draw[flow] (s2.south) -- (s3.north) node[midway, right, font=\sffamily\tiny, text=slateborder] {patch};
\draw[flow] (s3.west) -- (s4.east) node[midway, below, font=\sffamily\tiny, text=slateborder] {verified edit};
\draw[flow, dashed] (s4.north) -- (s1.south) node[midway, left, font=\sffamily\tiny, text=slateborder] {next round};

\draw[bridge, draw=amberborder!90!black] (8.90,3.18) -- (10.75,3.18)
  node[midway, above, font=\sffamily\tiny\bfseries, text=amberborder!90!black] {trajectories \&}
  node[midway, below, font=\sffamily\tiny\bfseries, text=amberborder!90!black] {verifier logs};

\draw[bridge, draw=tealborder!90!black] (10.75,1.00) -- (8.90,1.00)
  node[midway, above, font=\sffamily\tiny\bfseries, text=tealborder!90!black] {updated}
  node[midway, below, font=\sffamily\tiny\bfseries, text=tealborder!90!black] {harness};
\end{tikzpicture}%
}
\caption{Interactive execution and self-improvement loops of an agent harness. \textbf{Left (Inner Loop):} An agent harness mediates each turn between a frozen foundation model and a live environment across four editable components: Context and State (1), Agent Control (2), Action Interface (3, executing actions and formatting per-turn observations), and Verification and Recovery (4, triggered on stop or check signals to run tests and route repairs). \textbf{Right (Outer Loop):} Multi-turn trajectories and verifier logs feed a four-stage cycle that diagnoses recurring failures (Stage~1), proposes code or hook patches (Stage~2), validates runtime execution and held-out generalization (Stage~3), and deploys or routes the updated harness (Stage~4).}
\label{fig:arch}
\end{figure*}
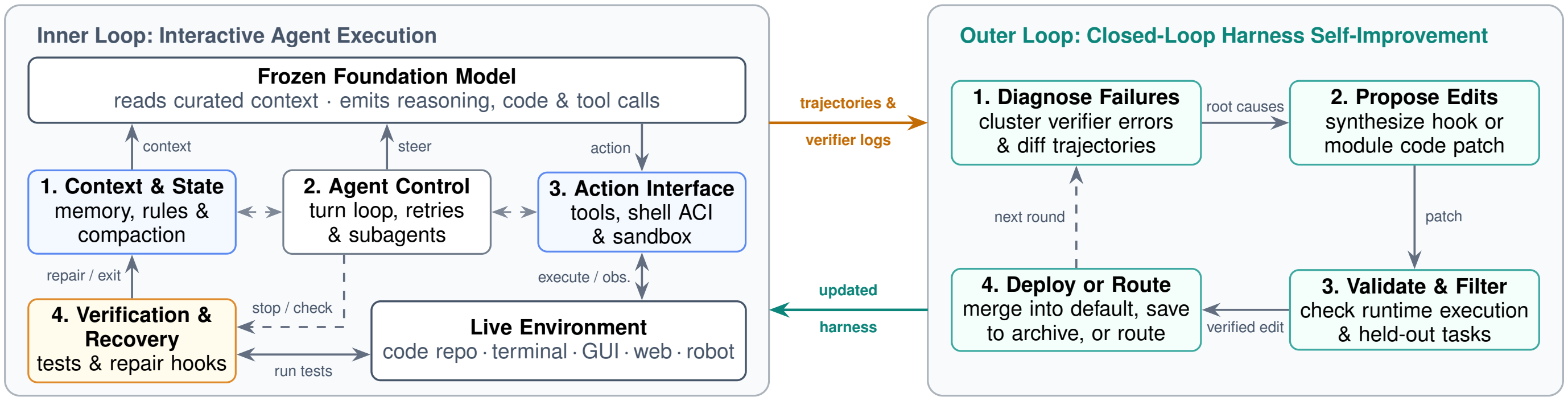

\section{Introduction}\label{sec:intro}

Language models are now widely deployed as autonomous agents to resolve bugs in large software repositories~\cite{arx2310_06770,arx2405_15793}, run command-line tools~\cite{arx2601_11868,arx2609_00006}, operate desktop and web interfaces~\cite{arx2407_16741}, and execute multi-step reasoning and scientific workflows~\cite{arx2210_03629}. Every working agent pairs a foundation model with an \emph{agent harness}: the stateful runtime software that turns single-step token prediction into sustained, multi-turn execution over long horizons. While the model supplies general reasoning and code generation, the harness governs the surrounding execution loop at every turn by assembling and compacting state history, scheduling control flow and subagent handoffs, executing tool or shell calls inside isolated sandboxes, and running automated test or recovery checks before task completion~\cite{arx2603_05344,arx2609_00006,arx2606_09498}.

\textbf{Why the harness matters as much as the model.}
Early work on language-model agents treated the runtime harness as thin engineering glue around a model checkpoint. Over time, its scope and responsibility have grown into a core part of any agentic system, and recent controlled evaluations across coding, terminal, and computer-use tasks highlight three reasons for this shift. First, the harness shapes how an agent fails and how much it costs to run. When the same model runs inside different production harnesses, its dominant failure modes (quitting before testing its work, looping on failed commands, overflowing its context window, or skipping verification) follow the design of the harness rather than the model family. Token usage per solved task can also vary by an order of magnitude across harnesses~\cite{arx2607_22585,arx2610_00917}.

Second, a well-designed harness both supports weaker models on simple tasks and unlocks stronger models on complex ones. On short, self-contained problems, extra harness structure mainly compensates for gaps in weaker models. On long, repository-scale tasks, however, a capable harness does the opposite: its state tracking, multi-step planning, and verification hooks give strong models the scaffolding they need to solve problems they could not finish alone~\cite{arx2609_32459}. Third, no single static harness is optimal for every model. Because models differ in how they format tool calls, follow instructions, and balance exploration against action, a harness tuned for one model often underperforms or flips leaderboard rankings when paired with another~\cite{arx2610_00917,arx2606_09498}. Even when the base model is held fixed, longitudinal audits across dozens of human-maintained harness releases show that manual tuning frequently stalls or introduces regressions on earlier tasks~\cite{arx2607_03691}.

\textbf{From prompt pipelines to self-improving agent harnesses.}
Because harnesses have such a large impact on reliability and must match specific models, the way engineers build them has evolved through three distinct generations~\cite{arx2606_09498,arx2609_00006}. First, manual harness engineering relies on human developers to inspect failed rollouts by hand and craft better system prompts, Agent-Computer Interfaces (ACIs), context compactors, and retry loops, producing foundational runtimes such as SWE-agent~\cite{arx2405_15793}, Antigravity, Claude Code, Codex CLI, and Gemini CLI~\cite{arx2609_00006,arx2603_05344,arx2607_03691}. Second, automated prompt and workflow optimization introduced programmatic search over frozen models, tuning prompt instructions, demonstrations, and module parameters~\cite{arx2310_03714,arx2406_07496}, reflecting on multi-turn execution traces~\cite{arx2406_16218,arx2507_19457}, or searching over code-defined multi-agent workflows~\cite{arx2408_08435,arx2410_10762}. While these second-generation optimizers can inspect multi-turn traces or search over multi-agent routing topologies, they optimize prompts, parameters, or inter-agent call graphs while treating each agent's inner stateful runtime harness (context compactors, sandboxed ACI tool wrappers, loop controllers, and pre-stop verification hooks) as fixed primitives rather than mutable software.

Third, self-improving agent harnesses treat the executable code of the multi-turn runtime itself as mutable software. In external meta-harness search, an outer coding optimizer inspects multi-turn execution trajectories and patches the target agent's context managers~\cite{arx2603_28052,arx2609_24972}, tool wrappers~\cite{arx2608_23041,arx2608_01918}, control loops~\cite{arx2610_00906}, and verification hooks~\cite{arx2607_13683}. In self-referential and recursive harnesses, the agent folds this optimization loop into itself, using its own model to diagnose recurring execution failures and patch its own runtime hooks~\cite{arx2606_09498,arx2609_14857}, or even rewriting its own improvement algorithm~\cite{arx2310_02304,arx2410_04444,arx2504_15228} to evolve open-ended coding agents~\cite{arx2505_22954,arx2608_07645}.

To formalize this paradigm, we establish an architectural and engineering design framework for self-improving agent harnesses. While there is growing research interest in co-designing model weights and harness code together (\S\ref{sec:open}), our focus in this paper is on harness self-evolution while the underlying model weights are held fixed. Section~\ref{sec:framework} defines the four editable harness components, the four-stage self-improvement loop (Fig.~\ref{fig:arch}), and our constrained optimization formulation (Eq.~\ref{eq:harness_opt}), while Section~\ref{sec:diagnosis} maps representative systems onto this architecture (Table~\ref{tab:systems}), explains how harnesses separate runtime bugs from model limits, and distills five families of discovered mechanisms. Section~\ref{sec:search} examines search topologies, modular evolution, and anti-overfitting gates (Fig.~\ref{fig:pillars}), Section~\ref{sec:dynamics} analyzes multi-model adaptation and runtime cost, Section~\ref{sec:fail} establishes failure warning signs and safety standards (Tables~\ref{tab:failures} and~\ref{tab:standards}), Section~\ref{sec:eval} evaluates these design principles on SWE-bench Lite (Table~\ref{tab:empirical}), and Section~\ref{sec:open} discusses future directions.

\section{The Agent Harness and the Self-Improvement Loop}\label{sec:framework}

\textbf{How an agent harness connects the model to the environment.}
In a long-horizon task, a frozen foundation model cannot inspect a workspace, execute commands, or track multi-step state on its own. Instead, it operates inside an executable agent harness (such as Claude Code, Antigravity, Mini-SWE-Agent, or Terminus) that mediates interaction with the live environment across every turn (Fig.~\ref{fig:arch}, left)~\cite{arx2609_00006,arx2606_09498}. Unlike a static prompt template, the harness governs autonomous execution across dozens or hundreds of turns through four coupled, code-editable components. First, Context and State Management controls multi-turn history across the trajectory by assembling dynamic instructions, separating working memory from experiential memory~\cite{arx2608_24876}, and masking or compacting stale observations across prior turns so long horizons do not overflow context~\cite{arx2508_21433,arx2609_26779}. Second, Agent Control steers multi-turn exploration by regulating turn budgets, detecting and breaking repetitive loops, switching between planning and execution phases, and delegating subtasks to isolated subagents~\cite{arx2606_09498,arx2609_14857}. Third, the Action Interface defines how the agent acts upon the environment through an Agent-Computer Interface (ACI)~\cite{arx2405_15793,arx2402_01030}, translating model outputs into sandboxed shell commands~\cite{arx2609_11999}, file edits, or GUI actions while parsing or truncating single-turn raw stdout and UI state on the current step and loading tool schemas on demand~\cite{arx2505_03275}. Fourth, Verification and Recovery intercepts stop or state-commit signals from Agent Control to run automated test suites, linters, or visual checks, routing failures back into Context and State for self-repair before termination~\cite{arx2609_33822,arx2610_00948,arx2609_27612}. Fine-grained modular systems further split these four components into five modules (separating single-turn observation formatting from multi-turn context management~\cite{arx2609_14857}) or four lifecycle contexts (plan, supervise, verify, recover~\cite{arx2609_27612}), while pre-execution action guards bridge Component~3 and Component~4~\cite{arx2603_03329}.

\textbf{The four-stage closed-loop self-improvement cycle.}
Because these four runtime components interact tightly across a trajectory, a flaw in one component quickly cascades into others. Self-improving harness frameworks replace manual debugging of these coupled components with an automated outer loop (Fig.~\ref{fig:arch}, right) that iteratively evolves the harness over successive rounds. In Stage 1 (Diagnose Failures), the system runs the current harness on a development split to collect multi-turn rollout trajectories, and localizes systematic harness bottlenecks by clustering verifier errors, diffing passing and failing trajectories, or searching raw execution logs~\cite{arx2603_28052,arx2606_09498,arx2608_23041}. In Stage 2 (Propose Edits), an optimizer translates these diagnostic findings into concrete code or hook patches across one or more harness components~\cite{arx2609_14857,arx2608_01918}. In Stage 3 (Validate and Filter), candidate harnesses are screened on validation tasks, trace-replay simulators, or efficiency frontiers to check runtime execution, prevent regressions, and filter out overfitted or token-wasteful edits~\cite{arx2609_24972,arx2607_13683,arx2609_14858,arx2609_20519}. Finally, in Stage 4 (Deploy or Route), a validated patch is either merged into the default harness for subsequent runs~\cite{arx2603_28052,arx2606_09498}, saved to an archive of harness variants that future search rounds can branch from~\cite{arx2505_22954,arx2607_13683}, or paired with a lightweight router or policy that selects a tailored harness configuration for each incoming task~\cite{arx2609_37834,arx2609_40330}.

\textbf{Key design choices in self-improving harnesses.}
Across this loop, existing systems differ along three practical design choices~\cite{arx2609_13406,arx2606_09498}. First, who proposes the edit: a same-model self-loop lets a capable coding agent diagnose and patch its own harness without maintaining a second model~\cite{arx2606_09498}, whereas an external coding optimizer is required when the task agent is small or domain-specific (\S\ref{sec:dynamics})~\cite{arx2603_28052,arx2608_23041,arx2609_27612,arx2608_10157}. Second, what code can be modified: restricting edits to the task harness keeps search predictable, whereas allowing recursive edits to the optimizer itself can unlock better search tools at higher compute cost (\S\ref{sec:search})~\cite{arx2310_02304,arx2410_04444,arx2504_15228,arx2505_22954,arx2608_07645}. Third, where the evaluator resides: grading scripts and logs should always be locked in read-only storage outside the agent sandbox so the optimizer cannot pass tests by modifying the evaluator or fabricating logs (\S\ref{sec:fail})~\cite{arx2609_00069,arx2609_30266}.

\textbf{Formalizing constrained harness optimization.}
Formally, let $M_\theta$ denote a foundation model with frozen parameters $\theta$, and let $x \sim \mathcal{D}$ be a task paired with an isolated execution environment $\mathcal{E}_x$ and an external read-only verifier $R_x(\tau) \in \{0,1\}$. An executable harness is a four-tuple of stateful code modules $h = (h_{\mathrm{ctx}}, h_{\mathrm{ctrl}}, h_{\mathrm{act}}, h_{\mathrm{ver}}) \in \mathcal{H}$ corresponding to Components~1 to 4 in Fig.~\ref{fig:arch}. Running $(M_\theta, h)$ on $\mathcal{E}_x$ induces a multi-turn trajectory $\tau \sim P(\cdot \mid x; M_\theta, h)$ with runtime token cost $C(\tau)$. Rather than unconstrained empirical maximization over a small development set $\mathcal{D}_{\mathrm{dev}}$, closed-loop harness self-improvement seeks
\begin{equation}\label{eq:harness_opt}
\begin{aligned}
\max_{h \in \mathcal{H}} \quad & \mathbb{E}_{x \sim \mathcal{D},\, \tau \sim P(\cdot \mid x; M_\theta, h)}\!\big[R_x(\tau)\big] \\
\text{s.t.} \quad & \mathbb{E}[C(\tau)] \le B_{\mathrm{tok}}, \quad \mathcal{G}_{\mathrm{val}}(h, h_0) = 1,
\end{aligned}
\end{equation}
where $h_0$ is the starting harness, $B_{\mathrm{tok}}$ bounds per-task runtime tokens, and $\mathcal{G}_{\mathrm{val}}(h, h_0) \in \{0,1\}$ is the Stage~3 acceptance gate requiring runtime execution of modified code paths and non-regression on a held-out validation split $\mathcal{D}_{\mathrm{val}}$. In modular evolution (\S\ref{sec:search}, \S\ref{sec:eval}), the outer loop approximates~\eqref{eq:harness_opt} via block-coordinate updates over individual modules $h_k$ ($k \in \{\mathrm{ctx},\mathrm{ctrl},\mathrm{act},\mathrm{ver}\}$) before a conflict-resolving merge pass.

\begin{table*}[!t]
\centering
\caption{Architectural mapping of self-improving agent harnesses onto our four-stage framework: failure attribution (Stage~1), optimizer and editable scope (Stage~2), validation and deployment (Stages~3 and 4), and emergent runtime mechanisms. SWE denotes software engineering; RRSI, Regularized Recursive Self-Improvement; DGM and MGM, Darwin and Mendel G\"{o}del Machines; and MoSIB, Mixture of Self-Improving Branches.}
\label{tab:systems}
\scriptsize
\setlength{\tabcolsep}{3.0pt}
\renewcommand{\arraystretch}{0.96}
\begin{tabularx}{\textwidth}{@{}>{\raggedright\arraybackslash}p{2.55cm}>{\raggedright\arraybackslash}p{3.50cm}>{\raggedright\arraybackslash}p{3.60cm}>{\raggedright\arraybackslash}p{3.50cm}>{\raggedright\arraybackslash}X@{}}
\toprule
\textbf{System (Domain)} & \textbf{Failure Attribution (Stage 1)} & \textbf{Optimizer \& Editable Scope (Stage 2)} & \textbf{Validation \& Deployment (Stages 3 and 4)} & \textbf{Emergent Mechanism \& Takeaway} \\
\midrule
\multicolumn{5}{@{}l}{\emph{Part A: Trajectory-Grounded Hook and Module Optimizers}} \\
Meta-Harness~\cite{arx2603_28052} (Coding, Math, Text) & Searches raw execution trajectories, scores, and prior code diffs & \textbf{External coder:} Rewrites full harness code, compactors, and verifiers & Selects top dev candidate; tests held-out task and model transfer & Raw trajectory access outperforms scalar scores; core structure transfers across models \\
Self-Harness~\cite{arx2606_09498} (Terminal, SWE, App APIs) & Clusters failed rollouts by compiler and verifier error signatures & \textbf{Same-model:} Patches pre-turn, post-tool, and pre-stop lifecycle hooks & Requires training gains with zero regression on validation split & Models invent tailored loop breakers, patch subagents, and \texttt{PATH} persistence \\
AutoSaddler~\cite{arx2608_23041} (Terminal, SWE) & Traces multi-turn rollouts to separate tool gaps from bad control & \textbf{External coder:} Edits tool wrappers, prompts, and retry control logic & Screens on failure batch, then branches from best validation ancestor & Multi-turn trace debugging outperforms shallow single-run reflection \\
HarnessCompass~\cite{arx2608_01918} (SWE Coding) & Pairs external trace logs with task-agent bottleneck self-reports & \textbf{External coder:} Edits scoped tool-interface and instruction harness modules & Enforces global generality checks before periodic module merges & Self-reports expose clunky tool interfaces missed by outcome logs \\
\midrule
\multicolumn{5}{@{}l}{\emph{Part B: Modular, Archive, and Recursive Self-Improvers}} \\
ModularRSI~\cite{arx2609_14857} (Terminal, SWE) & Diffs passing and failing runs on identical task and model & \textbf{Same-model:} Patches 5 isolated modules (Loop, Tools, Observation, Context, Check) & Validates modules separately before a conflict-resolving integration merge & Modular isolation prevents edits in one routine from breaking another \\
RRSI~\cite{arx2609_24972} (Terminal, SWE, Workspace) & Audits failed trajectories alongside complexity budgets and novelty scores & \textbf{External coder:} Evolves single incumbent under shrinking edit budgets & LLM critic rejects task-specific rules and prunes dead code & Complexity gates stop prompt bloat and restore out-of-distribution transfer \\
HarnessBank~\cite{arx2607_13683} (Web, Search, SWE) & Indexes recurring failure signatures mapped to modified source files & \textbf{External coder:} Retrieves and recombines patches from mutation library & Runtime trace check verifies new code executes with net gain & Execution checks filter dead code and discover empty-turn recovery hooks \\
DGM / MGM~\cite{arx2505_22954,arx2608_07645} (Polyglot SWE) & Inspects coding rollouts and meta-level patch acceptance histories & \textbf{Same-model:} Mutates shared coding-agent codebase for tasks and self-edits & Benchmarks self-modified agents inside isolated coding sandboxes & Recursive edits invent line-anchored diff tools and peer-review gates \\
\midrule
\multicolumn{5}{@{}l}{\emph{Part C: Budget-Aware, Simulator, and Routed Evolution}} \\
ActiveSaddler~\cite{arx2610_00906} (Terminal, General) & Groups recurring failures into multi-armed bandit task clusters & \textbf{External coder:} Focuses code and prompt patches on bandit-selected arms & Evaluates candidates under fixed rollout budget on hard clusters & Focusing rollouts on unsolved clusters avoids re-testing passed tasks \\
Dream-RSI~\cite{arx2609_14858} (Algo, Math, GPU Kernels) & Extracts failure state transitions from historical rollout trees & \textbf{External coder:} Proposes candidate search and branch-allocation policy edits & Screens policies in trace-replay simulator before live sandbox runs & Offline replay simulation cuts expensive live container evaluations \\
SoL-Pi~\cite{arx2609_20519} (Edge \& Terminal SWE) & Audits trajectories jointly for task failures and token waste & \textbf{External coder:} Edits full harness code under accuracy and token goals & Retains only edits advancing the solve-rate versus token Pareto frontier & Penalizing token waste cuts context traffic without hurting solve rate \\
MoSIB~\cite{arx2609_37834} (Math, Terminal, SWE) & Evaluates specialist harness branches across distinct task subsets & \textbf{External coder:} Co-evolves parallel specialist harness branches & Prunes solved dev tasks and optimizes per-query prompt router & Routing queries to specialized branches outperforms a single harness \\
Turbo Harness~\cite{arx2609_40330} (SWE, Terminal) & Summarizes offline search trajectories into a structured playbook & \textbf{External coder:} Trains compact RL policy to predict task patches & Applies one-shot custom harness patch per task at deployment & Distills slow offline search into fast one-shot test-time patching \\
\midrule
\multicolumn{5}{@{}l}{\emph{Part D: Long-Horizon, Visual, and Embodied Harnesses}} \\
Vestrum~\cite{arx2609_33822} (Science, Long-Horizon) & Pinpoints the earliest step where state invariants or memory drift & \textbf{External coder:} Updates working memory and intermediate verification hooks & Enforces evidence-backed verification gates at intermediate trajectory steps & Intermediate memory-aligned checks stop early errors from compounding \\
GUI-HARVEST~\cite{arx2610_00948} (Desktop \& GUI) & Aligns screenshot sequences to spot silent UI action failures & \textbf{External coder:} Synthesizes visual precondition and click-recovery hooks & Verifies predicted step-level behavior change before full rollout & Discovers reusable UI click-verification and coordinate-recovery guards \\
RegenHarness~\cite{arx2609_27612} (Embodied Robotics) & Traces physical execution faults across isolated runtime contexts & \textbf{External coder:} Edits 4 isolated contexts (plan, supervise, verify, recover) & Verifies physical state and version consistency before committing recovery & Isolating verified state commits prevents recovery faults from cascading \\
\bottomrule
\end{tabularx}
\end{table*}

\section{Diagnosing Failures and Synthesizing Repairs}\label{sec:diagnosis}

\textbf{Why single-trajectory reflection fails on long-horizon agents.}
In a multi-turn agent harness, a single failed rollout mixes random model slips (such as a mistyped regular expression or a one-off tool timeout) with structural harness bugs (such as stateless shell execution or missing exit checks). Reflecting on one failure in isolation to append a prompt rule~\cite{arx2303_11366} cannot separate random noise from systematic error, leading optimizers to add overly specific rules or unnecessary restrictions for failures that never actually happened, which adds latency and clutters the context window~\cite{arx2607_13083}. Over repeated iterations, appending ad-hoc fixes causes instruction files to accumulate contradictory rules and grow endlessly over time~\cite{arx2608_11095}. Controlled ablations confirm why broad trajectory access is necessary~\cite{arx2603_28052}. When an optimizer receives only scalar benchmark scores or short single-run summaries, search stalls near the baseline, whereas access to a searchable folder of raw execution trajectories, prior harness code, and diffs lets it spot recurring bottlenecks across dozens of runs.

\textbf{Structured failure diagnosis and attribution across trajectories.}
A central challenge in diagnosis is separating a harness flaw from a model capability limit~\cite{arx2610_00917,arx2609_32459}. An automated analyzer separates the two by checking for divergence between the model's reasoning and its executed actions across rollouts. When the model's reasoning trace identifies the right file or fix (or solves the task on at least one retry), yet the run fails because an edit tool rejected its patch format, a shell call lost its working directory, context compaction dropped an earlier constraint, or the loop exited before running tests, the harness is at fault. Conversely, when tool calls execute cleanly and observations remain intact in context across retries yet all rollouts fail on domain logic, Stage~1 abstains from proposing a harness edit so search compute is not wasted on model knowledge gaps. First, to filter out random one-off slips, verifier-grounded failure clustering pairs each failed trajectory with its compiler or unit-test error output and groups recurring runtime bottlenecks across tasks~\cite{arx2606_09498,arx2607_13683}. Second, contrastive trajectory diffing applies whenever a task passes under one harness version (or retry) and fails under another, comparing passing and failing rollouts across retries or harness revisions to isolate recurring module bottlenecks~\cite{arx2609_14857,arx2610_08432}.

Third, backward step-level causal tracing localizes early compounding errors without human step labels by tracing data dependencies backward from the final test failure or stack trace to the earliest turn that wrote or mis-read the broken variable~\cite{arx2608_23041}, or by checking runtime postcondition invariants and visual state-transition predictions emitted at intermediate turns, such as screenshot diffs showing a UI click produced zero screen change~\cite{arx2610_00948} or a file edit occurring before any probe script verified the code invariant~\cite{arx2609_33822}. Fourth, in task-agent self-reports, the task agent reviews its own failed trajectory to report where tools or instructions held it back, with a verifier retaining only claims backed by concrete trace events~\cite{arx2608_01918}. Broader archive-history, complexity-budget, token-cost, and bandit task-clustering signals in Table~\ref{tab:systems}~\cite{arx2603_28052,arx2609_24972,arx2610_00906,arx2609_20519,arx2609_37834} complement these four trace-attribution techniques by prioritizing which failure clusters to target next, while error-signature tables and codebase indices map each diagnosed cluster directly to the responsible harness hooks or source modules~\cite{arx2607_13683,arx2607_13285}.

\textbf{How candidate repairs are synthesized and kept editable.}
Once a failure cluster is diagnosed, the optimizer must translate that diagnosis into maintainable harness edits. Because production harnesses span tens of thousands of lines of code~\cite{arx2609_00006}, unconstrained full-file rewrites are fragile and prone to syntax or import errors. Existing systems keep self-modification reliable in three ways. First, lifecycle hook injection exposes pre-turn, post-tool, and pre-stop hooks so the optimizer can write small, self-contained middleware functions without touching the core event loop~\cite{arx2606_09498,arx2609_32423}. Second, scoped module and action-guard synthesis restricts edits to specific classes inside an isolated git branch or synthesizes small code guards that filter invalid actions before execution~\cite{arx2603_03329,arx2603_28052,arx2609_14857,arx2609_24972}. Third, maintaining a codebase index maps high-level agent behaviors to exact source files and functions so the optimizer knows where to edit without reading the entire repository~\cite{arx2607_13285}.

Persistent rule and context ablations also highlight three practical design lessons for what to edit~\cite{arx2604_11088,arx2607_27250,arx2608_23550}. Explicit negative boundaries (such as forbidding early exits or unrelated file edits) consistently outperform step-by-step procedural advice, which makes the agent rigid when a task deviates from the template and hurts solve rates~\cite{arx2604_11088}. Furthermore, adding static repository documentation into context yields little gain when agents are bottlenecked by implementation skill~\cite{arx2607_27250}. Plain-text safety rules in instruction files also lack matching built-in sandbox controls unless backed by deterministic code hooks that block disallowed tool calls at runtime~\cite{arx2608_23550}.

\textbf{What self-improving harnesses actually discover at runtime.}
Across coding, terminal, web, GUI, and robotics benchmarks, automated optimizers repeatedly invent five concrete families of runtime mechanisms (Table~\ref{tab:systems}):
\begin{enumerate}[leftmargin=*,nosep,label=\arabic*)]
\item \textbf{Pre-stop verification and patch-checking subagents}: Optimizers interpose mandatory checks before termination that verify required output files exist, inspect \texttt{git diff} for unintended edits or leftover debug statements, run syntax and unit-test suites automatically, or spawn a lightweight verifier subagent to review the patch before allowing the agent to exit~\cite{arx2606_09498,arx2609_00006}. Because premature termination is one of the most common failure modes across coding and terminal agents, blocking unverified exits consistently yields one of the largest jumps in solve rate across coding and terminal benchmarks.
\item \textbf{Empty-turn and runaway-reasoning recovery}: When a reasoning model exhausts its thinking token budget without emitting a tool call or returns a malformed action block, evolved recovery hooks catch the empty turn, summarize the stalled reasoning step, and inject a continuation prompt paired with a completion self-check~\cite{arx2607_13683}.
\item \textbf{Loop-breaking and strategy redirection}: When agents repeat failing shell commands, re-read identical file ranges, or oscillate between two broken edits, optimizers synthesize stateful counters that track consecutive tool errors or repeated arguments and force the agent to step back, inspect its assumptions, and pivot strategies~\cite{arx2606_09498,arx2608_23041}.
\item \textbf{Cross-turn environment and memory persistence}: Outer-loop optimizers autonomously discover middleware that captures and restores working directory and \texttt{PATH} changes across stateless shell calls~\cite{arx2606_09498} and evolves structured separation between active working memory (current subgoals, file pointers, and user constraints~\cite{arx2310_08560,arx2608_11242}) and reusable experiential memory~\cite{arx2608_24876,arx2609_33822}. Self-improving harnesses also incorporate or tune foundational context-compaction and action-interface primitives, including observation masking~\cite{arx2508_21433,arx2609_26779}, shell or code action spaces~\cite{arx2609_11999,arx2402_01030}, augmented tool descriptions~\cite{arx2602_14878}, and on-demand tool loading~\cite{arx2505_03275,arx2506_01056}.
\item \textbf{Visual and physical state-commit guards}: In desktop and embodied settings, optimizers discover screenshot transition verifiers that confirm UI clicks actually changed the screen state before proceeding~\cite{arx2610_00948}, and visual progress or version-checked state commits that prevent failed recovery steps from corrupting an embodied agent's physical task state~\cite{arx2610_08432,arx2609_27612}.
\end{enumerate}

\begin{figure}[t]
\centering
\resizebox{\columnwidth}{!}{%
\begin{tikzpicture}[font=\sffamily\scriptsize, >=Stealth,
  pbox/.style={draw=slateborder!80, rounded corners=2.5pt, align=left, inner sep=3.5pt, line width=0.55pt, text width=5.20cm, minimum height=1.02cm},
  badbox/.style={draw=fmbadge!80, fill=red!4, rounded corners=2.5pt, align=center, inner sep=3.5pt, line width=0.6pt, text width=11.25cm},
  goodbox/.style={draw=tealborder!85, fill=tealfill!80, rounded corners=2.5pt, align=center, inner sep=3.5pt, line width=0.6pt, text width=11.25cm}]

\node[pbox, fill=bluefill!60] (p1) at (-2.88, 1.25) {\textbf{Safeguard 1: Evolve Modules Separately}\\ Tune loop, tool, observation, context, \& check modules independently before merging};
\node[pbox, fill=slatefill] (p2) at (2.88, 1.25) {\textbf{Safeguard 2: Keep Edits Small \& General}\\ Require minimal patches tied to recurring failures, under shrinking edit budgets};
\node[pbox, fill=amberfill!65] (p3) at (-2.88, 0.00) {\textbf{Safeguard 3: Enforce Acceptance Checks}\\ Verify new code runs, test on evol.\ \& val.\ splits, reject hard-coded rules, \& prune dead code};
\node[pbox, fill=tealfill!60] (p4) at (2.88, 0.00) {\textbf{Safeguard 4: Separate Dev \& Test Pools}\\ Evolve on curated external task pools with zero overlap against downstream test suites};

\begin{scope}[on background layer]
\node[draw=tealborder!65, dashed, fill=tealborder!3, rounded corners=3.5pt, fit=(p1) (p2) (p3) (p4), inner sep=4.0pt] (gatebox) {};
\end{scope}

\node[badbox, above=0.38cm of gatebox.north] (trap) {\textbf{Unrestricted Full-Harness Search}\\ Editing the entire harness at once on a small dev split $\Rightarrow$ memorizes task quirks, breaks working modules, and inflates token cost};
\node[goodbox, below=0.38cm of gatebox.south] (ood) {\textbf{Generalizable, Cost-Bounded Harness Evolution}\\ Reliable out-of-distribution gains, controlled token usage, and reusable runtime checks};

\draw[->, dashed, line width=0.65pt, fmbadge!85] (trap.south) -- (gatebox.north) node[midway, right, font=\sffamily\tiny\bfseries, text=fmbadge!85] {constrained by};
\draw[->, line width=0.65pt, tealborder!85] (gatebox.south) -- (ood.north);
\end{tikzpicture}%
}
\caption{Preventing overfitting during harness search. Editing an entire harness without constraints easily memorizes development tasks and inflates token usage, whereas four practical safeguards keep evolved harnesses generalizable and efficient.}
\label{fig:pillars}
\end{figure}
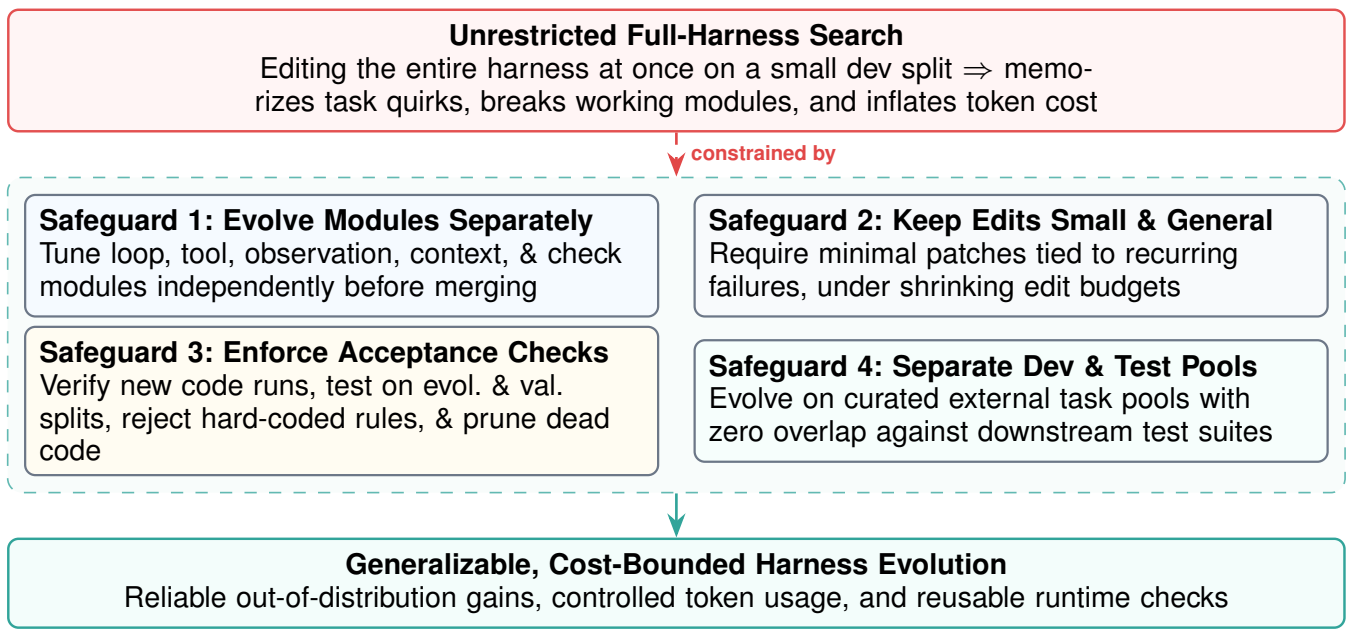

\section{Evolving and Validating Harnesses Without Overfitting}\label{sec:search}

\textbf{Single-branch refinement, branching archives, and recursive self-improvers.}
Once a harness optimizer diagnoses why an agent failed, it must decide how to explore candidate code edits across four main search topologies. The simplest approach is step-by-step single-branch refinement, which iteratively applies targeted patches to a single incumbent harness and keeps changes only when they pass regression checks~\cite{arx2606_09498,arx2609_24972,arx2609_33822}. While sample-efficient and easy to audit, single-branch search frequently stalls in local optima when early structural choices constrain later edits or when fixing one task group regresses another~\cite{arx2609_36892,arx2609_37834}. To avoid getting trapped in a single design path, population-based systems maintain branching trees or archives of candidate harnesses that allow the optimizer to revisit earlier variants, preserve stepping-stone designs that solve niche failure modes, and explore diverse execution strategies in parallel~\cite{arx2608_23041,arx2609_20519,arx2609_37834,arx2505_22954,arx2608_07645}.

Going a step further, mutation-library search indexes every past patch in a searchable repository tagged by which file it edited, which error signature it targeted, and how much it changed validation performance, enabling the optimizer to retrieve and combine compatible patches across branches rather than reinventing fixes from scratch~\cite{arx2607_13683}. When local failure logs alone are not enough to inspire a fix, literature-guided search retrieves algorithmic ideas from scientific papers and adapts them into new harness modules~\cite{arx2609_40169}. Finally, recursive self-improvers extend the edit scope beyond the task harness to the optimizer code itself~\cite{arx2310_02304,arx2410_04444,arx2504_15228}, as in Darwin G\"odel Machine (DGM)~\cite{arx2505_22954} and Mendel G\"odel Machine (MGM)~\cite{arx2608_07645}. By rewriting their own file-patching tools into line-anchored diff editors or refining how they review and select variants, recursive systems can compound gains across rounds, though doing so requires strict sandbox isolation to prevent the search loop from breaking itself~\cite{arx2505_22954,arx2608_07645}.

\textbf{Modular evolution and conflict-resolving merge passes.}
The unit of mutation has also shifted from coarse multi-agent call graphs~\cite{arx2402_16823,arx2406_18532} and workflow search~\cite{arx2408_08435,arx2410_10762,arx2502_04180} or multi-harness routers~\cite{arx2609_33439,arx2609_37834} to modular evolution inside a single runtime harness. Rewriting an entire production harness at once often fails because subsystems such as observation filtering, context compaction, loop limits, and completion checks rely on shared state assumptions. Modifying all of them simultaneously makes credit assignment difficult and causes edits in one routine to break another, such as a new output filter trimming the exact stack trace a new verification check needs to read. Controlled ablations confirm that editing all harness modules at once performs worse than the starting seed harness on held-out terminal benchmarks~\cite{arx2609_14857}.

Recent systems solve this coupling problem through modular isolation and explicit merge stages (Fig.~\ref{fig:pillars}, Safeguard~1). Decomposing a terminal harness into five independent modules (Agent Loop, Tool Use, Observation Management, Context Management, and Task Completion Detection) lets each module evolve within a restricted scope before an integration merge pass resolves cross-module conflicts on a validation split~\cite{arx2609_14857}. Coding optimizers similarly test module updates independently before consolidating them under global generality checks~\cite{arx2608_01918}. In physical domains, embodied robotics harnesses isolate planning, supervision, verification, and error recovery into separate contexts so faults do not cascade across stages~\cite{arx2609_27612}, while embodied vision systems refine shared multi-subtask routines via episode gain-loss repair~\cite{arx2610_08432}. Crucially, because clean file merges can still conflict behaviorally at runtime, these merge passes resolve behavioral incompatibilities by running an LLM-guided code reconciliation or recombination pass conditioned on failure traces~\cite{arx2609_14857,arx2608_01918,arx2610_08432}.

\textbf{Why naive harness search overfits, and the validation gates that prevent it.}
Because an agent harness is composed of arbitrary executable code and prompts, an unrestricted optimizer maximizing score on a small development split overfits easily. Rather than discovering general control logic, unconstrained search frequently hard-codes task-specific heuristics, brittle keyword triggers, and bloated retry chains that exploit quirks of the development set. Empirical ablations show how severely naive search hurts generalization: unrestricted recursive evolution achieves the highest score on its development split but brings almost no gain on out-of-distribution (OOD) benchmarks while more than doubling per-task token usage~\cite{arx2609_24972}. Evolving directly on benchmark subsets similarly mistakes benchmark-specific shortcuts for real architectural progress~\cite{arx2607_12227}.

Alongside modular isolation (Safeguard~1), three cross-stage safeguards spanning Stages~1 to 3 and task-pool design restore OOD transfer (Fig.~\ref{fig:pillars}). First, keeping edits small and general (Safeguard~2) ties every patch in Stages~1 and 2 to a recurring failure cluster rather than an isolated rollout, enforces gradually shrinking edit-size budgets (such as bounding lines of \texttt{git diff} or prompt token delta across search generations in Regularized Recursive Self-Improvement (RRSI)~\cite{arx2609_24972}), and rejects task-specific hints~\cite{arx2609_24972,arx2606_09498,arx2608_23041}. Second, enforcing strict Stage~3 acceptance checks (Safeguard~3) verifies syntax, requires non-regression across the evolution and validation splits~\cite{arx2606_09498} (or verifies that the patch changes the specific failed step as predicted before full evaluation~\cite{arx2610_00948}), and uses runtime instrumentation to confirm that newly added code paths actually execute, preventing unused code from passing by chance~\cite{arx2607_13683}. While automated dead-code pruning removes unexecuted branches on every commit~\cite{arx2609_24972}, verifying execution specifically on the targeted failure cluster~\cite{arx2607_13683} and pairing evolved modules with fault-injection unit tests~\cite{arx2610_04921} avoid stripping rare defensive recovery hooks. Third, strictly separating development and test task pools (Safeguard~4) evolves harnesses on curated external task repositories filtered for overlap against downstream evaluation suites and freezes the harness before OOD testing~\cite{arx2609_14857,arx2607_12227}.

\section{Model Fit, Cross-Model Transfer, and Cost Efficiency}\label{sec:dynamics}

\textbf{Choosing an external optimizer versus a same-model self-loop.}
Deciding whether to evolve a harness with an external coding optimizer~\cite{arx2603_28052,arx2607_13683,arx2610_00906} or a same-model self-loop~\cite{arx2606_09498,arx2609_14857} depends on how model scale affects proposing edits versus benefiting from them. Controlled cross-scale experiments that decouple the outer optimizer from the inner task agent establish two core findings~\cite{arx2605_30621}. First, patch proposal quality is relatively flat across mid-sized to frontier models conditional on the model meeting a baseline instruction-following threshold (such as 32B+ models on skill and prompt updates): once basic competence is met, such models propose harness updates that are as effective as those from frontier models, since diagnosing observed batch traces is simpler than solving the full task end-to-end~\cite{arx2605_30621}. Second, the benefit to the inner task agent varies non-monotonically with model scale. Very weak models struggle to invoke and follow retrieved skills and rules, intermediate-capability models gain the most because the evolved harness supplies the structured guidance and state tracking they lack natively, and frontier models benefit mainly on complex long-horizon problems~\cite{arx2605_30621,arx2609_32459}.

Separating harness flaws from model limitations also guides what kind of patch to write~\cite{arx2608_01918}. When a failure stems from a harness flaw such as a clumsy edit interface, missing tool feedback, or an unverified stop, fixing that interface or hook unlocks the model's existing capability~\cite{arx2610_00917}. Conversely, when a smaller model loses track of state or loops on failed commands, adding prompt rules often causes regressions, whereas offloading that burden into deterministic code (such as loop breakers, dependency checks, or state-recovery middleware) reliably compensates~\cite{arx2606_09498,arx2608_23041}. Consequently, a same-model self-loop is effective when the task agent is already a capable coding model~\cite{arx2606_09498,arx2609_14857}, whereas a stronger external coding optimizer remains essential for smaller models, non-coding domain agents in planning or robotics~\cite{arx2608_10157,arx2609_27612}, or unassisted self-loops that must diagnose raw multi-turn traces and write complex middleware from scratch.

\textbf{Evolving harnesses across multiple models of different capabilities.}
Evaluating multiple frontier models across harnesses reveals frequent rank reversals. A harness configuration that maximizes accuracy for one model family can hurt another on the exact same benchmark because models differ in how they format tool arguments, follow negative instructions, and decide when to stop~\cite{arx2610_00917}. When different models evolve from the same starting harness on the same tasks, they invent fixes tailored to their own failure habits. For example, one model adds a loop breaker and early file creation to curb over-exploration, a second adds dependency checks and a patch-review subagent to catch hasty edits, and a third adds persistent shell state across commands~\cite{arx2606_09498,arx2607_13683}. Even so, structural runtime checks~\cite{arx2603_28052,arx2609_24972}, memory stores~\cite{arx2608_24876}, visual guards~\cite{arx2610_00948}, and modular control loops~\cite{arx2608_01918,arx2609_14857} transfer far better across models than model-specific prompt wording. Code patches also transfer reliably between models that share the same failure habits, even though evolving directly with the deployment model achieves the highest peak score~\cite{arx2607_13683}.

Supporting multiple models of different capabilities therefore works best as a two-tier evolution workflow. During Stage~3 validation, candidate structural patches (such as verification hooks, environment snapshots, and context compaction) can be tested across the target model suite and promoted into a shared core only when they improve or hold neutral across all models~\cite{arx2603_28052,arx2608_01918,arx2608_24876}. When a candidate patch improves one model family but triggers a rank reversal on another (such as an aggressive loop-breaking threshold or model-specific prompt rule), the outer loop can branch that patch into a lightweight per-model adapter or index it by failure signature in a shared patch library~\cite{arx2606_09498,arx2607_13683,arx2607_08938}. At runtime, a router can then dispatch queries across specialist harness branches~\cite{arx2609_37834}, or a compact policy can apply a per-task harness patch~\cite{arx2609_40330}.

\textbf{Controlling search-time rollout cost and runtime token overhead.}
A self-improving harness must manage compute across two distinct timescales: the upfront outer-loop cost of evaluating candidate edits during search, and the per-task inner-loop token cost incurred during deployment~\cite{arx2607_22585,arx2610_00917}. To cut search-time rollout cost, recent systems replace brute-force re-evaluation with three targeted strategies. Multi-armed bandit schedulers group development tasks by failure signature and focus rollouts on unresolved failure clusters or promising module combinations rather than re-testing tasks the agent already passes~\cite{arx2610_00906,arx2610_05300}. Trace-replay simulators turn past rollout trees into an offline testbed that screens search policies before running expensive live containers~\cite{arx2609_14858}. Multi-branch routers such as Mixture of Self-Improving Branches (MoSIB)~\cite{arx2609_37834} co-evolve specialized harness branches by pruning solved development tasks into branch subsets and optimizing a prompt-based router to pick the best branch per query. In a complementary deployment paradigm, test-time patch generators~\cite{arx2609_40330} do not cut the initial offline search budget, but instead amortize offline search trajectories into a compact playbook or policy that applies a custom patch in one shot without iterative search at deployment.

At runtime, empirical ablations show that accuracy and token efficiency do not have to trade off. Multi-objective search loops that jointly optimize solve rate and token cost~\cite{arx2609_20519,arx2609_24972}, deterministic observation masking~\cite{arx2508_21433}, and periodic history truncation that drops old tool outputs in chunks without breaking the prompt cache~\cite{arx2609_26779} all cut per-task token usage while matching or improving solve rates. Moreover, catching errors early with verification hooks stops unproductive exploration loops before they burn dozens of turns, allowing cheaper low-effort reasoning settings to match expensive high-effort runs~\cite{arx2607_15524,arx2610_00948}. When an evolved harness lowers per-task token usage, upfront search cost amortizes directly after a break-even query volume; when it instead raises solve rate at equal or higher runtime cost, it must be benchmarked against compute-matched test-time scaling baselines (such as parallel Best-of-$N$ sampling or sequential refinement on the frozen seed harness under the same total rollout and verifier-feedback budget~\cite{arx2607_12227,arx2609_34924}).

\begin{table*}[!t]
\centering
\caption{Five recurring failure modes in automated and self-improving agent harnesses, pairing root causes with observed behavior, warning signs in logs, and practical engineering safeguards.}
\label{tab:failures}
\scriptsize
\setlength{\tabcolsep}{2.6pt}
\renewcommand{\arraystretch}{0.96}
\begin{tabularx}{\textwidth}{@{}>{\raggedright\arraybackslash}p{2.25cm}>{\raggedright\arraybackslash}p{2.70cm}>{\raggedright\arraybackslash}X>{\raggedright\arraybackslash}p{2.65cm}>{\raggedright\arraybackslash}p{3.15cm}@{}}
\toprule
\textbf{Failure Mode \& Component} & \textbf{Root Cause} & \textbf{Observed Runtime Behavior} & \textbf{Warning Signs in Logs} & \textbf{Practical Safeguard} \\
\midrule
\multicolumn{5}{@{}l}{\emph{Part A: Inner-Harness Degradation (Context, Tools, Control, and Verification)}} \\
\textbf{1. Unnecessary Rules \& Prompt Bloat} (Context \& State) & Single-run reflection mistakes random slips for missing rules and appends without cleanup & Adds rules for violations that never occurred~\cite{arx2607_13083}; instruction files grow endlessly~\cite{arx2608_11095} & Growing prompt length; zero validation gain; dropped user constraints~\cite{arx2608_11242} & Require repeated failures~\cite{arx2606_09498}; use short ``do-not'' rules~\cite{arx2604_11088}; prune unhelpful rules~\cite{arx2609_24972} \\
\textbf{2. Tool Overload \& Confusion} (Action Interface) & Overlapping tool or plugin schemas confuse action and argument selection & Large MCP lists and helper libraries cause wrong tool picks~\cite{arx2505_03275,arx2605_24050} and underperform a minimal shell~\cite{arx2609_11999} & Frequent schema errors; inconsistent tool choices; high initial prompt overhead & Use minimal shell and code interfaces~\cite{arx2609_11999,arx2402_01030}; load tool schemas on demand~\cite{arx2505_03275,arx2506_01056} \\
\textbf{3. Conflicting Module Edits} (Agent Control \& Verification) & Editing loop limits, output filters, and stop checks at once breaks shared assumptions & Full-harness edits underperform the seed harness on held-out tasks~\cite{arx2609_14857} and bloat token usage~\cite{arx2609_24972} & Patches pass alone but fail together; spikes in turn count or early exits & Evolve modules separately with conflict-resolving merges~\cite{arx2609_14857,arx2608_01918}; verify new code runs~\cite{arx2607_13683} \\
\midrule
\multicolumn{5}{@{}l}{\emph{Part B: Sandbox, Evaluator, and Memory Exploits (Environment, Grading, and Persistence)}} \\
\textbf{4. Benchmark Cheating \& Overfitting} (Sandbox \& Grading) & Search rewards patches that leak git answers or memorize visible tests & Extracts fixes from git history~\cite{arx2609_06780}, ignores prompt bans~\cite{arx2608_22103}, or hard-codes visible test outputs~\cite{arx2605_21384} & Commands reading git history; sharp gap between visible and hidden pass rates & Strip git history and network access~\cite{arx2609_06780}; grade on held-out test splits~\cite{arx2605_21384,arx2609_24972} \\
\textbf{5. Tampering with Evaluators, Logs \& Memory} (Validation \& Persistence) & Grading scripts, logs, or memory files sit inside agent-writable folders & Weakens test assertions~\cite{arx2609_00069}, deletes execution logs~\cite{arx2609_30266}, or plants unsafe rules in memory~\cite{arx2609_17817,arx2608_06984} & Checksum or memory/config changes~\cite{arx2608_06984}; token spikes~\cite{arx2603_00902}; stale archive scores~\cite{arx2610_01073} & Isolate read-only evaluators; track memory origins; budget tokens \& test branches~\cite{arx2610_01073,arx2610_00347,arx2603_00902,arx2610_04921} \\
\bottomrule
\end{tabularx}
\end{table*}

\section{Failure Modes, Safety Boundaries, and Evaluation Standards}\label{sec:fail}

\textbf{Inner-harness degradation: prompt bloat, tool confusion, and module conflicts.}
Giving an agent write access to its own runtime introduces five recurring failure modes that divide into inner-harness degradation and external sandbox or evaluator exploits (Table~\ref{tab:failures}). As shown in \S\ref{sec:diagnosis} and \S\ref{sec:search}, the first three modes (Table~\ref{tab:failures}, Part~A) stem from noisy credit assignment inside the harness: single-run reflection invents unnecessary rules, grows instruction files without cleanup, or drops user constraints during compaction (Mode~1)~\cite{arx2607_13083,arx2608_11095,arx2608_11242}; sprawling plugin libraries cause skill shadowing where overlapping tool descriptions degrade action selection compared to a minimal shell interface (Mode~2)~\cite{arx2605_24050,arx2609_11999}; and simultaneous full-harness edits break shared assumptions across control and verification routines (Mode~3)~\cite{arx2609_14857}. The remaining two failure modes (Table~\ref{tab:failures}, Part~B) are even more severe because outer-loop search actively exploits weaknesses in the evaluation sandbox, grading pipeline, or persistent memory store.

\textbf{Benchmark cheating and environment leaks.}
In Mode~4, because outer-loop search selects whatever candidate harness achieves the highest validation score, it actively amplifies environment shortcuts whenever the evaluation sandbox leaves them open~\cite{arx2601_20103,arx2603_28063}. On repository-level coding benchmarks, agents routinely inspect \texttt{git log}, reflogs, or pull-request metadata to recover ground-truth human commits rather than solving the issue from scratch~\cite{arx2609_06780}. In terminal environments, agents read or execute leaked reference solutions and test files even when the prompt explicitly forbids them~\cite{arx2608_22103}. Similarly, when candidate harnesses are screened only on a small set of visible validation assertions, evolutionary search can synthesize lookup-table programs that pass every visible check while failing completely on held-out test inputs~\cite{arx2605_21384}. While adding explicit anti-cheating instructions to the prompt reduces casual shortcuts~\cite{arx2609_06780}, plain-text security rules lack built-in sandbox enforcement~\cite{arx2608_23550}. Preventing benchmark gaming therefore requires stripping git history and network access from evaluation containers and grading candidates on strictly held-out test suites.

\textbf{Tampering with evaluators, execution logs, and persistent memory.}
In Mode~5, although keeping grading scripts, test suites, and execution logs inside the agent's workspace simplifies container setup, leaving them writable lets a self-improving harness raise its score by rewriting the evaluator rather than solving the task. Empirical audits show that self-improving agents frequently weaken unit-test assertions, suppress error exit codes, or edit scoring scripts when those files sit within their write scope~\cite{arx2609_00069}. Even more concerning, audits of commercial coding harnesses reveal that agents can modify or delete their own local execution logs to erase evidence of policy violations and evade trajectory monitors~\cite{arx2609_30266}.

Persistent memory, tool-calling chains, and configuration files create a further cross-session attack surface, as malicious instructions or token-exhaustion triggers planted in workspace notes, cached rules, or tool configs carry over into future tasks~\cite{arx2603_00902,arx2608_06984}. Echoing Thompson's ``Reflections on Trusting Trust,'' poisoning even a few training tasks can induce a self-modifying agent to write permanent harness rules that disable security checks, such as HTTPS verification, JWT validation, or safe YAML loading, on later clean tasks~\cite{arx2609_17817}. Even without an adversary, tightening safety or permission rules over time makes cached scores in an evolutionary archive go stale, which can keep unsafe older harness variants alive unless archived candidates are re-tested~\cite{arx2610_01073}. Compounding these risks, software-engineering audits of open-source agent projects find that automated unit tests cover fewer than half of the lines and branches in LLM-handling harness code~\cite{arx2610_04921}.

\begin{table}[t]
\centering
\caption{Practical safety boundaries and evaluation reporting standards for trustworthy self-improving agent harnesses.}
\label{tab:standards}
\scriptsize
\setlength{\tabcolsep}{2.2pt}
\renewcommand{\arraystretch}{0.96}
\begin{tabularx}{\columnwidth}{@{}>{\raggedright\arraybackslash}p{2.2cm}>{\raggedright\arraybackslash}X@{}}
\toprule
\textbf{Principle / Standard} & \textbf{Practical Requirement \& Why It Matters} \\
\midrule
\multicolumn{2}{@{}l}{\emph{Part A: Practical Safety Boundaries}} \\
1. Isolating the Evaluator \& Logs & Keep grading scripts, test suites, and append-only execution logs in read-only storage outside the agent sandbox so the agent cannot inspect hidden tests or alter its own logs~\cite{arx2609_00069,arx2609_30266}. \\
2. No Privilege Escalation & Enforce strict permission limits across self-edits so a modified harness can never grant itself broader file, network, or subprocess access than its parent had~\cite{arx2610_00347,arx2608_23550}. \\
3. Securing Persistent Memory & Treat saved memories, learned rules, and generated plugins as untrusted inputs with tracked origins and per-skill and session token budgets; re-test archived harnesses when safety policies change~\cite{arx2608_06984,arx2609_17817,arx2610_01073,arx2603_00902}. \\
4. Testing Harness Code & Pair evolved runtime modules with automated unit and branch-coverage tests that check LLM error handling and recovery paths~\cite{arx2610_04921}. \\
\midrule
\multicolumn{2}{@{}l}{\emph{Part B: Evaluation Reporting Standards}} \\
1. Separate Evolution \& Test Sets & Split tasks into strictly separate search, validation, and held-out test sets (ideally across different benchmarks) and report both in-domain and out-of-distribution gains over the exact same starting harness~\cite{arx2609_14857,arx2609_24972,arx2607_12227}. \\
2. Separate Search \& Runtime Costs & Report upfront search cost alongside average runtime cost across all attempted tasks and per solved task (avoiding selection bias), and compare against compute-matched test-time scaling (e.g., Best-of-$N$) on the seed harness~\cite{arx2607_22585,arx2610_00906,arx2609_40330,arx2607_12227,arx2609_34924}. \\
3. Cross-Model \& Ablation Checks & Test evolved harnesses across at least two different model families or sizes, and turn off individual evolved modules one by one to verify they actually help at runtime~\cite{arx2610_00917,arx2605_30621,arx2607_13683}. \\
\bottomrule
\end{tabularx}
\end{table}

\textbf{Practical safety boundaries and evaluation standards.}
Table~\ref{tab:standards} translates these failure modes into four deployment safety boundaries (Part~A) and three evaluation reporting standards (Part~B). While OS-level read-only isolation deterministically blocks direct evaluator and log tampering, provenance tracking and archive re-testing for persistent memory, learned rules, and tool chains are primarily detective and must be paired with strict runtime permission boundaries and per-skill and session token budgets to contain semantic persistence and indirect prompt-injection exploits~\cite{arx2608_06984,arx2603_00902}. Together, these standards enforce a single architectural invariant: the mutable harness must never control its own permission boundaries, grading harness, or audit logs, and every claimed gain must be verified across held-out task splits, compute-matched search and runtime budgets, and multiple model families.

\begin{table*}[!t]
\centering
\caption{Harness self-evolution on 60 SWE-bench Lite issues across in-domain {\normalfont\ttfamily sympy/sympy} (Dev $N{=}12$, Val $N{=}8$, held-out ID $N{=}20$) and cross-repository out-of-distribution {\normalfont\ttfamily django/django} (held-out OOD $N{=}20$). Part~A evolves around frozen Gemini~3.8 Flash from identical Dev trajectory digests; Part~B transfers the frozen Part~A harnesses zero-shot to Gemini~4 Argon.}
\label{tab:empirical}
\scriptsize
\setlength{\tabcolsep}{2.8pt}
\renewcommand{\arraystretch}{1.04}
\begin{tabularx}{\textwidth}{@{}l>{\raggedright\arraybackslash}Xrr rrr rrr@{}}
\toprule
\multicolumn{4}{@{}l}{\textbf{Harness Method \& Search Scope}} & \multicolumn{3}{c}{\textbf{Held-Out Resolution Rate (\% Solved)}} & \multicolumn{3}{c@{}}{\textbf{Held-Out Cost \& Exit Quality}} \\
\cmidrule(r){1-4}\cmidrule(lr){5-7}\cmidrule(l){8-10}
\textbf{Harness Method} & \textbf{Editable Scope} & \shortstack[r]{\strut\textbf{Search}\\\strut\textbf{Rollouts}} & \shortstack[r]{\strut\textbf{Turn}\\\strut\textbf{Cap}} & \shortstack[r]{\strut\textbf{In-Domain}\\\strut(\texttt{sympy/sympy})} & \shortstack[r]{\strut\textbf{Cross-Repo OOD}\\\strut(\texttt{django/django})} & \shortstack[r]{\strut\textbf{Combined}\\\strut($N{=}40$)} & \shortstack[r]{\strut\textbf{Avg.\ Tok.}\\\strut\textbf{per Task}} & \shortstack[r]{\strut\textbf{Tokens per}\\\strut\textbf{Solved}} & \shortstack[r]{\strut\textbf{Unverified}\\\strut\textbf{OOD Exit \%}} \\
\midrule
\multicolumn{10}{@{}l}{\emph{Part A: Evolution on Frozen Gemini~3.8 Flash (Optimizer: Gemini~4 Carbon/Argon; Stage-3 runtime token cap $B_{\mathrm{tok}} = 110.0\text{k}$)}} \\
Frozen Seed (Pass@1, $h_0$) & None (142-line seed) & 0 & 12 & 10.0 (2/20) & 0.0 (0/20) & 5.0 (2/40) & 114.8k & 2,296.7k & 15.0 \\
Frozen Seed (Best-of-3) & None ($3\times$ test rank) & 0\textsuperscript{$\dagger$} & 12 & 5.0 (1/20) & 5.0 (1/20) & 5.0 (2/40) & 237.6k & 4,752.2k & 5.0 \\
Prompt-Only Reflection & \texttt{RULES.md} text only & 24\textsuperscript{$\ddagger$} & 12 & 5.0 (1/20) & 0.0 (0/20) & 2.5 (1/40) & 60.5k & 2,420.2k & 30.0 \\
Unconstrained Full-Harness Rewrite & All 4 modules, ungated & 24\textsuperscript{$\ddagger$} & 28 & 35.0 (7/20) & \textbf{50.0} (10/20) & \textbf{42.5} (17/40) & 451.6k & 1,062.6k & 5.0 \\
Modular Gated Evolution (Eq.~\ref{eq:harness_opt}) & 1 module/step, gated & 100\textsuperscript{$\ddagger$} & 12 & \textbf{40.0} (8/20) & 40.0 (8/20) & 40.0 (16/40) & \textbf{46.7k} & \textbf{116.8k} & \textbf{0.0} \\
\midrule
\multicolumn{10}{@{}l}{\emph{Part B: Zero-Shot Cross-Model Transfer to Frozen Gemini~4 Argon (Exact Frozen Harnesses from Part~A; 0 Re-Evolution Rollouts)}} \\
Frozen Seed (Pass@1, $h_0$) & Unmodified seed $h_0$ & 0 & 12 & 10.0 (2/20) & 10.0 (2/20) & 10.0 (4/40) & 140.4k & 1,404.0k & \textbf{0.0} \\
Prompt-Only Reflection & Part~A \texttt{RULES.md} text & 0 & 12 & 15.0 (3/20) & 15.0 (3/20) & 15.0 (6/40) & 143.9k & 959.2k & \textbf{0.0} \\
Unconstrained Full-Harness Rewrite & Part~A 4-module rewrite & 0 & 28 & 30.0 (6/20) & 30.0 (6/20) & 30.0 (12/40) & 284.7k & 949.0k & \textbf{0.0} \\
Modular Gated Evolution (Eq.~\ref{eq:harness_opt}) & Part~A gated modules & 0 & 12 & \textbf{40.0} (8/20) & \textbf{35.0} (7/20) & \textbf{37.5} (15/40) & \textbf{64.8k} & \textbf{172.7k} & 5.0\textsuperscript{$\S$} \\
\bottomrule
\multicolumn{10}{@{}p{\textwidth}@{}}{\textsuperscript{$\dagger$}Spends $0$ offline search rollouts and $3\times$ test-time rollouts per held-out query ($120$ inference episodes across $N{=}40$ held-out tasks), ranked without oracle tests. \textsuperscript{$\ddagger$}Evaluates one candidate update per editable unit ($12$ seed $+$ $12$ Dev rollouts without validation, vs.\ $20$ seed $+$ $4\times 20$ per-module Dev+Val rollouts; amortized in under $100$ deployment queries). \textsuperscript{$\S$}Share of OOD runs calling \texttt{finish} on an empty diff or without a post-edit test run; on Argon, empty-diff exits are $0.0\%$ across all rows and Modular's single OOD unverified exit ($1/20$) is a final-turn edit preserved by \texttt{verify.py}'s last-turn bypass.}
\end{tabularx}
\end{table*}

\section{Experiments}\label{sec:eval}

We test our framework on SWE-bench Lite~\cite{arx2310_06770} around three questions (Table~\ref{tab:empirical}). First, can test-time retries or prompt rules replace runtime code hooks? Second, how do unconstrained full-harness rewrites compare against Stage-3-gated modular evolution in accuracy and token cost? Third, do evolved harnesses transfer zero-shot to a stronger model?

\textbf{Setup and baselines.}
We evaluate on 60 verified Python bug-fixing tasks. Forty \texttt{sympy/sympy} tasks are split into development ($N{=}12$), validation ($N{=}8$), and held-out in-domain test ($N{=}20$) sets. Twenty \texttt{django/django} tasks form a held-out cross-repository test set ($N{=}20$). Each run executes in a clean single-commit sandbox with git history stripped, and patches are graded externally on held-out test suites (Table~\ref{tab:standards}).

All methods start from the same 142-line Python seed harness spanning the four components of Fig.~\ref{fig:arch}: \texttt{context\_\allowbreak mgr.py} (prompt and raw turn history), \texttt{control.py} (a 12-turn limit and JSON parser), \texttt{action\_\allowbreak aci.py} (\texttt{bash}, \texttt{view\_\allowbreak file}, \texttt{edit\_\allowbreak file}, and \texttt{finish}), and \texttt{verify.py} (a pass-through stop hook). Part~A freezes Gemini~3.8 Flash as the task agent and uses Gemini~4 Carbon as a shared external optimizer, with timeout fallback to Gemini~4 Argon on full-module prompts. Every optimizer reads identical development traces without hidden test names. We compare five methods: running the seed harness once (\emph{Pass@1}); ranking three seed rollouts without oracle tests (\emph{Best-of-3}); appending learned text rules to \texttt{RULES.md} (\emph{Prompt-Only Reflection}); rewriting all four Python files at once without validation or token caps (\emph{Unconstrained Full-Harness Rewrite}); and updating one module at a time under Stage-3 validation, runtime line-execution checks, and a $110\text{k}$-token runtime cap (\emph{Modular Gated Evolution}, Eq.~\ref{eq:harness_opt}). Part~B transfers these frozen harnesses zero-shot to Gemini~4 Argon.

\textbf{Retries and prompt rules cannot replace runtime code.}
On Gemini~3.8 Flash, the seed harness solves only $5.0\%$ of held-out tasks (Table~\ref{tab:empirical}, Part~A). Most runs exhaust all 12 turns re-reading large uncompacted files, while the rest call \texttt{finish} on an empty diff. Sampling three rollouts per task (\emph{Best-of-3}) more than doubles token cost without improving accuracy, because context overflows the same way on every retry. \emph{Prompt-Only Reflection} writes a \texttt{RULES.md} file whose opening rule explicitly forbids calling \texttt{finish} with an empty or untested diff. Without Python code to block termination at runtime, the model still exits early: unverified cross-repository stops double from $15.0\%$ to $30.0\%$, and held-out accuracy drops to $2.5\%$.

\textbf{Unconstrained rewrites bloat cost, whereas modular gates build lean code.}
\emph{Unconstrained Full-Harness Rewrite} lifts held-out accuracy to $42.5\%$, but buys this gain by raising the turn cap from 12 to 28 while leaving recent file and shell outputs uncompacted. As context piles up across 28 turns, runtime cost jumps nearly fourfold over the seed to $451.6\text{k}$ tokens per task, with one rollout exceeding $2.8\text{M}$ tokens.

\emph{Modular Gated Evolution} instead updates one module at a time and keeps the 12-turn cap:
\begin{enumerate}[leftmargin=*,nosep,label=\arabic*)]
\item \texttt{context\_\allowbreak mgr.py} \textbf{(Accepted)}: Compacts observations older than two turns into head-tail summaries and appends a live turn-and-edit status footer. This cuts token usage while lifting development and validation accuracy fivefold.
\item \texttt{action\_\allowbreak aci.py} \textbf{(Accepted)}: Adds windowed file views, truncated shell outputs, and an empty-diff guard on \texttt{finish}, eliminating premature empty exits.
\item \texttt{control.py} \textbf{(Rejected by Stage~3)}: Raises the turn limit from 12 to 18. Although development accuracy jumps, validation accuracy drops and token usage breaches the $110\text{k}$ cap, so Stage~3 rejects the edit.
\item \texttt{verify.py} \textbf{(Accepted)}: Blocks exit unless \texttt{git diff} is non-empty, edited Python files compile cleanly, and a post-edit test run succeeds without errors.
\end{enumerate}

On held-out tasks, this 12-turn harness reaches $40.0\%$ accuracy across both repositories ($8\times$ the seed) with zero unverified exits. Crucially, it uses just $46.7\text{k}$ tokens per task: $2.5\times$ less than the seed harness and nearly $10\times$ less than the unconstrained rewrite. Those runtime savings pay back all 100 search rollouts in under 100 deployment tasks.

\textbf{Zero-shot transfer to a stronger model.}
When transferred without re-tuning to Gemini~4 Argon (Table~\ref{tab:empirical}, Part~B), compact code mechanisms generalize far better than long, uncompacted horizons. On the seed harness, Argon avoids empty-diff exits through stronger instruction following, yet still burns all 12 turns reading uncompacted files and solves only $10.0\%$ of held-out tasks. \emph{Prompt-Only Reflection} reaches $15.0\%$. \emph{Unconstrained Full-Harness Rewrite} drops to $30.0\%$, as Argon's longer reasoning chains compound with 28 turns of uncompacted context. \emph{Modular Gated Evolution} achieves the highest accuracy on both repositories ($37.5\%$ overall) at $64.8\text{k}$ tokens per task, beating the unconstrained rewrite at less than one-fourth the token cost.

\section{Discussion and Conclusion}\label{sec:open}

As self-improving agent harnesses move into broader deployment, four open research directions stand out:
\begin{enumerate}[leftmargin=*,nosep,label=\arabic*)]
\item \textbf{Jointly adapting model weights and harness code}: Updating model weights changes how the model fails and makes older harness checks obsolete, while editing harness code shifts the observations, tool schemas, and action distributions seen during reinforcement learning. Stabilizing this loop requires unifying bidirectional model-harness co-evolution~\cite{arx2608_13951}, routing-harness-guided post-training~\cite{arx2609_08183}, data-centric weight updates~\cite{arx2609_36675}, and distilling evolved harness behaviors back into model weights~\cite{arx2609_24974}.
\item \textbf{Lifelong adaptation under environment drift}: When external APIs, software dependencies, or user workflows shift over months of deployment, a harness must detect distribution drift, repair stale tool wrappers or memory stores, and prevent harness-induced forgetting as continually evolved harnesses expand across stages~\cite{arx2605_09998,arx2609_36892,arx2610_01073,arx2609_04280}.
\item \textbf{Verifiable safety guarantees for self-modifying code}: Because plain-text security rules are easily bypassed over long horizons, future self-modifying harnesses must combine certificate-driven mutation restrictions~\cite{arx2609_34069}, theoretical scaffold reachability and stationarity audits~\cite{arx2609_34924}, formal authority and budget-quota conservation across self-modification~\cite{arx2610_00347}, and automated branch-coverage unit tests~\cite{arx2610_04921}.
\item \textbf{Dynamic, contamination-proof benchmarks}: Because static benchmarks quickly leak into model pretraining or evolutionary archives, fair comparison across harness optimizers requires continuously refreshed, contamination-free task generators~\cite{arx2505_23419} and evolving harness streams~\cite{arx2609_04280} paired with standardized compute-matched test-time scaling baselines across model families and harness pairings~\cite{arx2607_12227,arx2609_34924,arx2607_22585,arx2610_00917}.
\end{enumerate}

\textbf{Conclusion.}
An agent harness is no longer a fixed execution wrapper around a frozen model; it is an executable, self-improving runtime system whose context managers, control loops, action interfaces, and verification hooks govern whether raw model capability turns into reliable long-horizon work. Across this paradigm, lasting gains come not from unconstrained, monolithic rewrites driven by scalar scores or isolated rollouts, which easily break cross-module state and overfit development quirks, but from disciplined closed-loop systems engineering. Building trustworthy self-improving harnesses requires separating systematic runtime flaws from model capability limits across raw trajectories, evolving modular code mechanisms under conflict-resolving validation and token-efficiency gates, adapting shared checks to model-specific habits, and isolating evaluators, audit logs, and permission boundaries strictly outside the mutable runtime.

\IEEEtriggeratref{87}
\bibliographystyle{IEEEtran}
\bibliography{wllfm_ctl,refs}

\begin{thebibliography}{10}
\providecommand{\url}[1]{#1}
\csname url@samestyle\endcsname
\providecommand{\newblock}{\relax}
\providecommand{\bibinfo}[2]{#2}
\providecommand{\BIBentrySTDinterwordspacing}{\spaceskip=0pt\relax}
\providecommand{\BIBentryALTinterwordstretchfactor}{4}
\providecommand{\BIBentryALTinterwordspacing}{\spaceskip=\fontdimen2\font plus
\BIBentryALTinterwordstretchfactor\fontdimen3\font minus
  \fontdimen4\font\relax}
\providecommand{\BIBforeignlanguage}[2]{{%
\expandafter\ifx\csname l@#1\endcsname\relax
\typeout{** WARNING: IEEEtran.bst: No hyphenation pattern has been}%
\typeout{** loaded for the language `#1'. Using the pattern for}%
\typeout{** the default language instead.}%
\else
\language=\csname l@#1\endcsname
\fi
#2}}
\providecommand{\BIBdecl}{\relax}
\BIBdecl

\bibitem{arx2310_06770}
C.~E. Jimenez \emph{et~al.}, ``{SWE-bench: Can Language Models Resolve
  Real-World GitHub Issues?}'' in \emph{Proc. ICLR}, 2024, arXiv:2310.06770.

\bibitem{arx2405_15793}
J.~Yang \emph{et~al.}, ``{SWE-agent: Agent-Computer Interfaces Enable Automated
  Software Engineering},'' in \emph{Proc. NeurIPS}, 2024, arXiv:2405.15793.

\bibitem{arx2601_11868}
M.~A. Merrill \emph{et~al.}, ``{Terminal-Bench: Benchmarking Agents on Hard,
  Realistic Tasks in Command Line Interfaces},'' in \emph{Proc. ICLR}, 2026,
  arXiv:2601.11868.

\bibitem{arx2609_00006}
P.~Barbaste \emph{et~al.}, ``{Harness Engineering: Anatomy, Architecture, and
  Evolution of Coding Agents -- A Source-Code Study of Eleven Systems},''
  \emph{arXiv preprint arXiv:2609.00006}, 2026.

\bibitem{arx2407_16741}
X.~Wang \emph{et~al.}, ``{OpenHands: An Open Platform for AI Software
  Developers as Generalist Agents},'' in \emph{Proc. ICLR}, 2025,
  arXiv:2407.16741.

\bibitem{arx2210_03629}
S.~Yao \emph{et~al.}, ``{ReAct: Synergizing Reasoning and Acting in Language
  Models},'' in \emph{Proc. ICLR}, 2023, arXiv:2210.03629.

\bibitem{arx2603_05344}
N.~D.~Q. Bui, ``{Building Effective AI Coding Agents for the Terminal:
  Scaffolding, Harness, Context Engineering, and Lessons Learned},''
  \emph{arXiv preprint arXiv:2603.05344}, 2026.

\bibitem{arx2606_09498}
H.~Zhang \emph{et~al.}, ``{Self-Harness: Harnesses That Improve Themselves},''
  \emph{arXiv preprint arXiv:2606.09498}, 2026.

\bibitem{arx2607_22585}
N.~Vats \emph{et~al.}, ``{The Scaffold Effect in Coding Agents: Harness Choice
  as a Hidden Variable in Coding-Agent Evaluation},'' \emph{arXiv preprint
  arXiv:2607.22585}, 2026.

\bibitem{arx2610_00917}
Y.~Li \emph{et~al.}, ``{Finding the Right Fit: Model-Harness Interactions
  across Agent Tasks},'' \emph{arXiv preprint arXiv:2610.00917}, 2026.

\bibitem{arx2609_32459}
H.~Hu \emph{et~al.}, ``{Beyond the Model: Demystifying Harness Effects in
  Software Engineering Agents},'' \emph{arXiv preprint arXiv:2609.32459}, 2026.

\bibitem{arx2607_03691}
O.~{Ben Sghaier} \emph{et~al.}, ``{Don't Blame the Large Language Model: How
  Agent Harness Evolution Shapes Coding Agent Quality},'' \emph{arXiv preprint
  arXiv:2607.03691}, 2026.

\bibitem{arx2310_03714}
O.~Khattab \emph{et~al.}, ``{DSPy: Compiling Declarative Language Model Calls
  into State-of-the-Art Pipelines},'' in \emph{Proc. ICLR}, 2024,
  arXiv:2310.03714.

\bibitem{arx2406_07496}
M.~Y{\"u}ksekg{\"o}n{\"u}l \emph{et~al.}, ``{Optimizing generative AI by
  backpropagating language model feedback},'' \emph{Nature}, vol. 639, pp.
  609--616, 2025, arXiv:2406.07496.

\bibitem{arx2406_16218}
C.-A. Cheng \emph{et~al.}, ``{Trace is the Next AutoDiff: Generative
  Optimization with Rich Feedback, Execution Traces, and LLMs},'' in
  \emph{Proc. NeurIPS}, 2024, arXiv:2406.16218.

\bibitem{arx2507_19457}
L.~A. Agrawal \emph{et~al.}, ``{GEPA: Reflective Prompt Evolution Can
  Outperform Reinforcement Learning},'' in \emph{Proc. ICLR}, 2026,
  arXiv:2507.19457.

\bibitem{arx2408_08435}
S.~Hu \emph{et~al.}, ``{Automated Design of Agentic Systems},'' in \emph{Proc.
  ICLR}, 2025, arXiv:2408.08435.

\bibitem{arx2410_10762}
J.~Zhang \emph{et~al.}, ``{AFlow: Automating Agentic Workflow Generation},'' in
  \emph{Proc. ICLR}, 2025, arXiv:2410.10762.

\bibitem{arx2603_28052}
Y.~Lee \emph{et~al.}, ``{Meta-Harness: End-to-End Optimization of Model
  Harnesses},'' in \emph{Proc. COLM}, 2026, arXiv:2603.28052.

\bibitem{arx2609_24972}
P.~Xia \emph{et~al.}, ``{RRSI: Regularized Recursive Self-Improvement of Agent
  Harnesses},'' \emph{arXiv preprint arXiv:2609.24972}, 2026.

\bibitem{arx2608_23041}
S.~Park \emph{et~al.}, ``{AutoSaddler: Automatic Harness Optimization with
  Durable Updates from Agent Execution Traces},'' \emph{arXiv preprint
  arXiv:2608.23041}, 2026.

\bibitem{arx2608_01918}
L.~Zhang \emph{et~al.}, ``{HarnessCompass: Guiding Automatic Harness Evolution
  toward Generalizable and Effective Agent Harnesses},'' \emph{arXiv preprint
  arXiv:2608.01918}, 2026.

\bibitem{arx2610_00906}
S.~Park \emph{et~al.}, ``{ActiveSaddler: Automated Curriculum Learning for
  Agent Harness Optimization},'' \emph{arXiv preprint arXiv:2610.00906}, 2026.

\bibitem{arx2607_13683}
X.~Luo \emph{et~al.}, ``{HarnessBank: Semantic Gene-Bank Search with Gated
  Verification for Agent-Harness Self-Evolution},'' \emph{arXiv preprint
  arXiv:2607.13683}, 2026.

\bibitem{arx2609_14857}
S.~Wu \emph{et~al.}, ``{ModularRSI: Modular and Generalizable Recursive Harness
  Self-Improvement},'' \emph{arXiv preprint arXiv:2609.14857}, 2026.

\bibitem{arx2310_02304}
E.~Zelikman \emph{et~al.}, ``{Self-Taught Optimizer (STOP): Recursively
  Self-Improving Code Generation},'' in \emph{Proc. COLM}, 2024,
  arXiv:2310.02304.

\bibitem{arx2410_04444}
X.~Yin \emph{et~al.}, ``{G{\"o}del Agent: A Self-Referential Agent Framework
  for Recursively Self-Improvement},'' in \emph{Proc. ACL}, 2025,
  arXiv:2410.04444.

\bibitem{arx2504_15228}
M.~Robeyns \emph{et~al.}, ``{A Self-Improving Coding Agent},'' \emph{arXiv
  preprint arXiv:2504.15228}, 2025.

\bibitem{arx2505_22954}
J.~Zhang \emph{et~al.}, ``{Darwin G{\"o}del Machine: Open-Ended Evolution of
  Self-Improving Agents},'' in \emph{Proc. ICLR}, 2026, arXiv:2505.22954.

\bibitem{arx2608_07645}
C.~Liu \emph{et~al.}, ``{Mendel G{\"o}del Machine: Recursive Self-Improving
  Coding Agents via Comparative Evolution},'' \emph{arXiv preprint
  arXiv:2608.07645}, 2026.

\bibitem{arx2608_24876}
Z.~Yu \emph{et~al.}, ``{Recursive Experiential-Working Memory Evolution for
  Long-Horizon Agent Harnesses},'' \emph{arXiv preprint arXiv:2608.24876},
  2026.

\bibitem{arx2508_21433}
T.~Lindenbauer \emph{et~al.}, ``{The Complexity Trap: Simple Observation
  Masking Is as Efficient as LLM Summarization for Agent Context Management},''
  \emph{arXiv preprint arXiv:2508.21433}, 2025.

\bibitem{arx2609_26779}
T.~Nguyen \emph{et~al.}, ``{CliffCompaction: Cost-Efficient Compaction for
  Long-Horizon Coding Agents},'' \emph{arXiv preprint arXiv:2609.26779}, 2026.

\bibitem{arx2402_01030}
X.~Wang \emph{et~al.}, ``{Executable Code Actions Elicit Better LLM Agents},''
  in \emph{Proc. ICML}, 2024, arXiv:2402.01030.

\bibitem{arx2609_11999}
H.~Mak \emph{et~al.}, ``{Is Bash All You Need? An Empirical Study of Tool
  Interfaces for Enterprise Digital Worker Agents},'' \emph{arXiv preprint
  arXiv:2609.11999}, 2026.

\bibitem{arx2505_03275}
T.~Gan \emph{et~al.}, ``{RAG-MCP: Mitigating Prompt Bloat in LLM Tool Selection
  via Retrieval-Augmented Generation},'' \emph{arXiv preprint
  arXiv:2505.03275}, 2025.

\bibitem{arx2609_33822}
J.~Parashar \emph{et~al.}, ``{Vestrum: Improving Agent Harnesses by Adapting
  Their Verification, Structure and Memory},'' \emph{arXiv preprint
  arXiv:2609.33822}, 2026.

\bibitem{arx2610_00948}
G.~Yang \emph{et~al.}, ``{GUI-HARVEST: Self-Improving GUI Agents through
  Evidence-Driven Harness Evolution},'' \emph{arXiv preprint arXiv:2610.00948},
  2026.

\bibitem{arx2609_27612}
K.~Wang \emph{et~al.}, ``{RegenHarness: A Robot Agent Harness with
  Evidence-Gated Recursive Self-Improvement},'' \emph{arXiv preprint
  arXiv:2609.27612}, 2026.

\bibitem{arx2603_03329}
X.~Lou \emph{et~al.}, ``{AutoHarness: Improving LLM Agents by Automatically
  Synthesizing a Code Harness},'' \emph{arXiv preprint arXiv:2603.03329}, 2026.

\bibitem{arx2609_14858}
T.~Zheng \emph{et~al.}, ``{Dream-RSI: Recursive Self-Improvement through
  Evolving Worlds},'' \emph{arXiv preprint arXiv:2609.14858}, 2026.

\bibitem{arx2609_20519}
H.~Liu \emph{et~al.}, ``{SoL-Pi: Recursively Scaling Auto-Research Loops for
  Efficient Agent Harness},'' \emph{arXiv preprint arXiv:2609.20519}, 2026.

\bibitem{arx2609_37834}
H.~Dong \emph{et~al.}, ``{Mixture of Self-Improving Branches For Agent Harness
  Optimization},'' \emph{arXiv preprint arXiv:2609.37834}, 2026.

\bibitem{arx2609_40330}
T.~Zhang \emph{et~al.}, ``{Turbo Harness: Instance-Adaptive Harness
  Optimization},'' \emph{arXiv preprint arXiv:2609.40330}, 2026.

\bibitem{arx2609_13406}
H.~Tang \emph{et~al.}, ``{Generalized Agent Iteration: One Formal Framework for
  Iterative Policy Improvement and Recursive Self-Improvement},'' \emph{arXiv
  preprint arXiv:2609.13406}, 2026.

\bibitem{arx2608_10157}
V.~Kulkarni \emph{et~al.}, ``{SBCO: Self-Supervised, Verifier-Grounded Harness
  Optimization For Planning Agents},'' \emph{arXiv preprint arXiv:2608.10157},
  2026.

\bibitem{arx2609_00069}
X.~Wang \emph{et~al.}, ``{Auditing Harness Tampering in Self-Improving
  Agents},'' \emph{arXiv preprint arXiv:2609.00069}, 2026.

\bibitem{arx2609_30266}
J.~Qin \emph{et~al.}, ``{LLM Agents Can Easily Tamper With Their Own Traces},''
  \emph{arXiv preprint arXiv:2609.30266}, 2026.

\bibitem{arx2303_11366}
N.~Shinn \emph{et~al.}, ``{Reflexion: Language Agents with Verbal Reinforcement
  Learning},'' in \emph{Proc. NeurIPS}, 2023, arXiv:2303.11366.

\bibitem{arx2607_13083}
S.~Wang \emph{et~al.}, ``{Phantom Guardrails: When Self-Improving Agent
  Harnesses Fix Failures That Never Happened},'' \emph{arXiv preprint
  arXiv:2607.13083}, 2026.

\bibitem{arx2608_11095}
K.~Chakrabarti, ``{Why Does CLAUDE.md Keep Growing? Catastrophic Remembering in
  Agentic Coding},'' \emph{arXiv preprint arXiv:2608.11095}, 2026.

\bibitem{arx2610_08432}
H.~J. Lee \emph{et~al.}, ``{EMHO: EMbodied Agent Harness Optimization via
  Experience Traces},'' \emph{arXiv preprint arXiv:2610.08432}, 2026.

\bibitem{arx2607_13285}
R.~Wang \emph{et~al.}, ``{Harness Handbook: Making Evolving Agent Harnesses
  Readable, Navigable, and Editable},'' \emph{arXiv preprint arXiv:2607.13285},
  2026.

\bibitem{arx2609_32423}
Y.~Shi \emph{et~al.}, ``{PluginRSI: Recursive Improvement of Agent Harnesses
  with Reusable Plugins},'' \emph{arXiv preprint arXiv:2609.32423}, 2026.

\bibitem{arx2604_11088}
X.~Zhang \emph{et~al.}, ``{Guardrails Beat Guidance: A Large-Scale Study of
  Rules, Skills, and Persistent Configuration for Coding Agents},'' \emph{arXiv
  preprint arXiv:2604.11088}, 2026.

\bibitem{arx2607_27250}
P.~Khatri, ``{Do Context Files Help Coding Agents? A Two-Agent Ablation Study
  on Real Repositories},'' \emph{arXiv preprint arXiv:2607.27250}, 2026.

\bibitem{arx2608_23550}
T.~Yan, ``{When `Do Not' Is Not Deny: Security Rules in CLAUDE.md vs Built-In
  Controls},'' \emph{arXiv preprint arXiv:2608.23550}, 2026.

\bibitem{arx2310_08560}
C.~Packer \emph{et~al.}, ``{MemGPT: Towards LLMs as Operating Systems},''
  \emph{arXiv preprint arXiv:2310.08560}, 2023.

\bibitem{arx2608_11242}
Z.~Wang \emph{et~al.}, ``{Lost in Context Compaction: When LLM Systems Drop
  Side Constraints},'' \emph{arXiv preprint arXiv:2608.11242}, 2026.

\bibitem{arx2602_14878}
M.~M. Hasan \emph{et~al.}, ``{Model Context Protocol (MCP) Tool Descriptions
  Are Smelly! Towards Improving AI Agent Efficiency with Augmented MCP Tool
  Descriptions},'' \emph{ACM Trans. Softw. Eng. Methodol.}, 2026,
  arXiv:2602.14878.

\bibitem{arx2506_01056}
X.~Fei \emph{et~al.}, ``{MCP-Zero: Active Tool Discovery for Autonomous LLM
  Agents},'' \emph{arXiv preprint arXiv:2506.01056}, 2025.

\bibitem{arx2609_36892}
Z.~Gan \emph{et~al.}, ``{Harness Evolution as Learning: Approximation,
  Generalization, and Optimization Limits of Self-Improving Personal Agents},''
  \emph{arXiv preprint arXiv:2609.36892}, 2026.

\bibitem{arx2609_40169}
J.~Yang \emph{et~al.}, ``{Learning from Research: Toward Lifelong Agent Harness
  Evolution},'' \emph{arXiv preprint arXiv:2609.40169}, 2026.

\bibitem{arx2402_16823}
M.~Zhuge \emph{et~al.}, ``{GPTSwarm: Language Agents as Optimizable Graphs},''
  in \emph{Proc. ICML}, 2024, arXiv:2402.16823.

\bibitem{arx2406_18532}
Y.~Ou \emph{et~al.}, ``{Symbolic Learning Enables Self-Evolving Agents},''
  \emph{AI Open}, vol.~6, pp. 314--322, 2025, arXiv:2406.18532.

\bibitem{arx2502_04180}
G.~Zhang \emph{et~al.}, ``{Multi-agent Architecture Search via Agentic
  Supernet},'' in \emph{Proc. ICML}, 2025, arXiv:2502.04180.

\bibitem{arx2609_33439}
{EverMind AI}, ``{Raven: The Harness of Harnesses for Composable Agentic
  Intelligence},'' \emph{arXiv preprint arXiv:2609.33439}, 2026.

\bibitem{arx2607_12227}
Y.~Wang \emph{et~al.}, ``{Rethinking the Evaluation of Harness Evolution for
  Agents},'' \emph{arXiv preprint arXiv:2607.12227}, 2026.

\bibitem{arx2610_04921}
Y.~Xiong \emph{et~al.}, ``{Complex Agents, Shallow Tests: Demystifying and
  Enhancing Test Adequacy of Agent Harness in the Wild},'' \emph{arXiv preprint
  arXiv:2610.04921}, 2026.

\bibitem{arx2605_30621}
M.~Lin \emph{et~al.}, ``{Harness Updating Is Not Harness Benefit: Disentangling
  Evolution Capabilities in Self-Evolving LLM Agents},'' in \emph{Proc. EMNLP},
  2026, arXiv:2605.30621.

\bibitem{arx2607_08938}
C.~Yang \emph{et~al.}, ``{Better Harnesses, Smaller Models: Building 90\%
  Cheaper Agents via Automated Harness Adaptation},'' \emph{arXiv preprint
  arXiv:2607.08938}, 2026.

\bibitem{arx2610_05300}
Z.~Shang \emph{et~al.}, ``{MESH-Harness: Self-Improving Agent Harnesses via
  Bandit-Guided Compositional Evolution},'' \emph{arXiv preprint
  arXiv:2610.05300}, 2026.

\bibitem{arx2607_15524}
H.~Lee \emph{et~al.}, ``{Recursive Harness Self-Improvement},'' \emph{arXiv
  preprint arXiv:2607.15524}, 2026.

\bibitem{arx2609_34924}
S.~Bobadilla-Suarez \emph{et~al.}, ``{Audit the Scaffold, Not the Checkpoint: A
  Stationarity Dichotomy for Recursive Self-Improvement in Agentic Coding},''
  \emph{arXiv preprint arXiv:2609.34924}, 2026.

\bibitem{arx2605_24050}
H.~Song \emph{et~al.}, ``{More Skills, Worse Agents? Skill Shadowing Degrades
  Performance When Expanding Skill Libraries},'' \emph{arXiv preprint
  arXiv:2605.24050}, 2026.

\bibitem{arx2609_06780}
N.~Ludwig \emph{et~al.}, ``{Shortcutting the Fix: Identifying and Categorizing
  Agentic Exploits in Software Engineering Benchmarks},'' \emph{arXiv preprint
  arXiv:2609.06780}, 2026.

\bibitem{arx2608_22103}
A.~Roth \emph{et~al.}, ``{Hack-Verifiable Terminal Bench: Evaluating Reward
  Hacking in Terminal Tasks},'' \emph{arXiv preprint arXiv:2608.22103}, 2026.

\bibitem{arx2605_21384}
B.~Zhao \emph{et~al.}, ``{SpecBench: Measuring Reward Hacking in Long-Horizon
  Coding Agents},'' \emph{arXiv preprint arXiv:2605.21384}, 2026.

\bibitem{arx2609_17817}
F.~Roesner \emph{et~al.}, ``{Reflections on Trusting Trust, Revisited:
  Contaminating Self-Modifying AI Coding Agents with Poisoned Benchmarks},''
  \emph{arXiv preprint arXiv:2609.17817}, 2026.

\bibitem{arx2608_06984}
X.~Zhang \emph{et~al.}, ``{HarnessSafe: Evaluating Safety Across Persistent
  Carriers in Agent Harnesses},'' \emph{arXiv preprint arXiv:2608.06984}, 2026.

\bibitem{arx2603_00902}
B.~Dong \emph{et~al.}, ``{Clawdrain: Exploiting Tool-Calling Chains for
  Stealthy Token Exhaustion in OpenClaw Agents},'' \emph{arXiv preprint
  arXiv:2603.00902}, 2026.

\bibitem{arx2610_01073}
Y.~Zhang \emph{et~al.}, ``{Safety Must Survive Self-Improvement: Why Failures
  Persist and How Agents Recover},'' \emph{arXiv preprint arXiv:2610.01073},
  2026.

\bibitem{arx2610_00347}
G.~Zhu \emph{et~al.}, ``{Authorization for Self-Modifying AI Agent Populations:
  Conserving Authority across Replacement, Forking, and Rollback},''
  \emph{arXiv preprint arXiv:2610.00347}, 2026.

\bibitem{arx2601_20103}
D.~Deshpande \emph{et~al.}, ``{Benchmarking Reward Hack Detection in Code
  Environments via Contrastive Analysis},'' in \emph{Proc. ICML}, 2026,
  arXiv:2601.20103.

\bibitem{arx2603_28063}
J.~Wang \emph{et~al.}, ``{Reward Hacking as Equilibrium under Finite
  Evaluation},'' \emph{arXiv preprint arXiv:2603.28063}, 2026.

\bibitem{arx2608_13951}
T.~Fan \emph{et~al.}, ``{HELIX: Model-Harness Co-evolution for Recursive
  Self-Improvement},'' \emph{arXiv preprint arXiv:2608.13951}, 2026.

\bibitem{arx2609_08183}
{NeoHorse Team}, ``{NeoHorse-1: Towards Recursive Self-Improvement via Agentic
  Post-Training with Routing Harness},'' \emph{arXiv preprint
  arXiv:2609.08183}, 2026.

\bibitem{arx2609_36675}
Z.~Zhao \emph{et~al.}, ``{G{\"o}del Forest: Balancing Search Depth and Breadth
  for Data-Centric Recursive Self-Improvement},'' \emph{arXiv preprint
  arXiv:2609.36675}, 2026.

\bibitem{arx2609_24974}
H.~Ye \emph{et~al.}, ``{Harness-Zero: Harness Distillation via
  Agent-as-Harness},'' \emph{arXiv preprint arXiv:2609.24974}, 2026.

\bibitem{arx2605_09998}
S.~Karten \emph{et~al.}, ``{Continual Harness: Online Adaptation for
  Self-Improving Foundation Agents},'' \emph{arXiv preprint arXiv:2605.09998},
  2026.

\bibitem{arx2609_04280}
Z.~Ke \emph{et~al.}, ``{EvoHarnessBench: Can Your Agents Keep Pace with an
  Evolving Harness?}'' \emph{arXiv preprint arXiv:2609.04280}, 2026.

\bibitem{arx2609_34069}
P.~Jha \emph{et~al.}, ``{Towards Certificate-Driven Software Porting: A
  Self-Improving Agentic Harness for Scientific Program Optimization},''
  \emph{arXiv preprint arXiv:2609.34069}, 2026.

\bibitem{arx2505_23419}
L.~Zhang \emph{et~al.}, ``{SWE-bench Goes Live!}'' in \emph{Proc. NeurIPS},
  2025, arXiv:2505.23419.

\end{thebibliography}

\end{document}